\documentclass[screen]{acmart}

\acmDOI{1111111.1111111}

\usepackage[english]{babel}

\usepackage{amsmath}
\usepackage{graphicx}
\usepackage{setspace}
\usepackage{amsmath}
\usepackage{amsthm}
\usepackage[ruled]{algorithm2e}
\usepackage{algpseudocode}
\usepackage{microtype}
\usepackage{enumitem}
\usepackage{arydshln}
\usepackage{bm}
\usepackage[export]{adjustbox}
\usepackage{booktabs}
\usepackage{caption}
\usepackage{subcaption}
\usepackage{setspace}
\usepackage{mathtools}
\usepackage{cleveref}
\usepackage{tikz}
\usepackage{float}

\newcommand{\mcal}[1]{\mathcal{#1}}
\newcommand{\mbb}[1]{\mathbb{#1}}

\newcommand{\acc}{\ensuremath{\mathrm{acc}}}
\newcommand{\conf}{\ensuremath{\mathrm{conf}}}

\usetikzlibrary{shapes.geometric, arrows}

\tikzstyle{rect} = [rectangle, 
rounded corners, 
minimum width=7cm,
minimum height=1cm,
text centered,
draw=black,
fill=red!30
]

\tikzstyle{rect_wrap} = [rectangle, 
rounded corners, 
minimum width=3cm,
minimum height=1cm,
text width=4cm,
text centered,
draw=black,
fill=red!30
]

\tikzstyle{rect_medium} = [rectangle, 
rounded corners, 
minimum width=3cm,
minimum height=1cm,
text width=5cm,
text centered,
draw=black,
fill=red!30
]

\tikzstyle{rect_large} = [rectangle, 
rounded corners, 
minimum width=5cm,
minimum height=1cm,
text width=5cm,
text centered,
draw=black,
fill=red!30
]

\tikzstyle{rect_larger} = [rectangle, 
rounded corners, 
minimum width=7cm,
minimum height=1cm,
text width=7cm,
text centered,
draw=black,
fill=red!30
]

\definecolor{row0_color}{RGB}{142, 164, 255}
\definecolor{row1_color}{RGB}{200, 255, 178}
\definecolor{row2_color}{RGB}{251, 189, 255}
\definecolor{row3_color}{RGB}{251, 255, 186}

\tikzstyle{arrow} = [thick, -, >=stealth]

\begin{document}

\title{A Survey on Uncertainty Quantification of Large Language Models: Taxonomy, Open Research Challenges, and Future Directions}

\author{Ola Shorinwa}
\email{shoa@princeton.edu}
\author{Zhiting Mei}
\email{maymei@princeton.edu}
\author{Justin Lidard}
\email{jlidard@princeton.edu}
\author{Allen Z. Ren}
\email{allen.ren@princeton.edu}
\author{Anirudha \mbox{Majumdar}}
\email{ani.majumdar@princeton.edu}
\affiliation{%
  \institution{\\ \mbox{Princeton} \mbox{University}}
  \city{Princeton}
  \state{NJ}
  \country{USA}
}

\renewcommand{\shortauthors}{}

\begin{abstract}
    The remarkable performance of large language models (LLMs) in content generation, coding, and common-sense reasoning has spurred widespread integration into many facets of society. However, integration of LLMs raises valid questions on their reliability and trustworthiness, given their propensity to generate hallucinations: plausible, factually-incorrect responses, which are expressed with striking confidence. 
Previous work has shown that hallucinations and other non-factual responses generated by LLMs can be detected by examining the uncertainty of the LLM in its response to the pertinent prompt, driving significant research efforts devoted to quantifying the uncertainty of LLMs. This survey seeks to provide an extensive review of existing uncertainty quantification methods for LLMs, identifying their salient features, along with their strengths and weaknesses. We present existing methods within a relevant taxonomy, unifying ostensibly disparate methods to aid understanding of the state of the art.
Furthermore, we highlight applications of uncertainty quantification methods for LLMs, spanning chatbot and textual applications to embodied artificial intelligence applications in robotics. We conclude with open research challenges in uncertainty quantification of LLMs, seeking to motivate future research.
\end{abstract}



\maketitle

\section{Introduction}
\label{sec:introduction}
Large language models have demonstrated remarkable language generation capabilities, surpassing average human performance on many benchmarks including math, reasoning, and coding \cite{achiam2023gpt, anthropic2024claude, brown2020language, touvron2023llama, dubey2024llama, chiang2023vicuna}. For example, recent (multi-modal) large language models were shown to achieve impressive scores, e.g., in the $90\%$ percentile, on simulated Law School Admission Test (LSAT) exams, the American Mathematics Competition (AMC) contests, the Multistate Bar Exam, and the Graduate Record Exam (GRE) General Test, outperforming a majority of test takers \cite{katz2024gpt, achiam2023gpt, anthropic2024claude}. Likewise, LLMs have advanced the state of the art in machine translation, text summarization, and question-and-answer tasks. However, LLMs also tend to produce plausible, factually-incorrect responses to their input prompts, termed \emph{hallucinations} \cite{lee2018hallucinations}. In some scenarios, the hallucinated response is overtly incorrect; however, in many cases, the factuality of the LLM response is harder to discern, posing significant risk as a user might falsely assume factuality of the response, which can result in devastating consequences, especially when safety is of paramount importance. As a result, hallucinations pose a notable danger to the safe, widespread adoption of LLMs.

\begin{figure}[th]
    \centering
    \includegraphics[width=0.5\columnwidth]{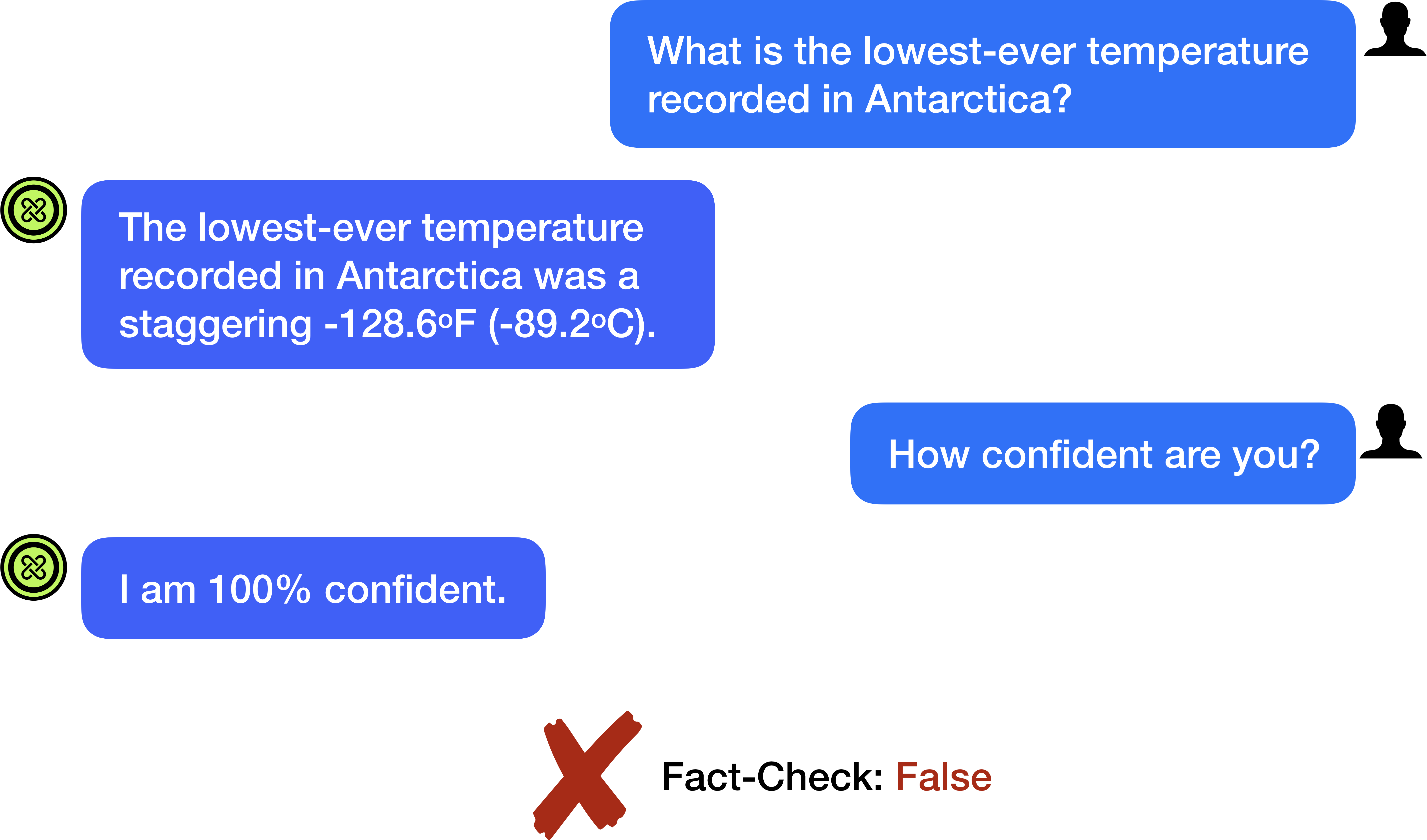}
    \caption{A user asks an LLM the question: \emph{What is the lowest-ever temperature recorded in Antarctica?}; in response, the LLM answers definitively. Afterwards, the user asks the LLM how confident the LLM is. Although the LLM states that it is ``100\% confident," the LLM's response fails to pass a fact-check test. Confidence scores provided by LLMs are generally miscalibrated. UQ methods seek to provide calibrated estimates of the confidence of LLMs in their interaction with users.
    }
    \Description[A user asks an LLM the question: \emph{What is the lowest-ever temperature recorded in Antarctica?}; in response, the LLM answers definitively. Afterwards, the user asks the LLM how confident the LLM is. Although the LLM states that it is ``100\% confident," the LLM's response fails to pass a fact-check test. Confidence scores provided by LLMs are generally miscalibrated. UQ methods seek to provide calibrated estimates of the confidence of LLMs in their interaction with users.]{A user asks an LLM the question: \emph{What is the lowest-ever temperature recorded in Antarctica?}; in response, the LLM answers definitively. Afterwards, the user asks the LLM how confident the LLM is. Although the LLM states that it is ``100\% confident," the LLM's response fails to pass a fact-check test. Confidence scores provided by LLMs are generally miscalibrated. UQ methods seek to provide calibrated estimates of the confidence of LLMs in their interaction with users.
    }
    \label{fig:uq_for_llms}
\end{figure}

To ensure the trustworthiness of LLMs, substantial research has been devoted to examining the mechanisms behind hallucinations in LLMs \cite{lee2018hallucinations, chen2023hallucination, azamfirei2023large, xu2024hallucination, ji2023survey}, detecting its occurrence, identifying potential causes, and proposing mitigating actions. However, even in the absence of hallucinations, LLMs are susceptible to doubt when given prompts at the boundary of their knowledge base. In these situations, prior work has shown that LLMs fail to accurately convey their uncertainty to a user, either implicitly or explicitly, unlike typical humans \cite{liu2023trustworthy, alkaissi2023artificial}. In fact, LLMs tend to be overconfident even when they should be uncertain about the factuality of their response \cite{xiong2023can, groot2024overconfidence}. We provide an example in \Cref{fig:uq_for_llms}, where an LLM is asked: ``What is the lowest-ever temperature recorded in Antarctica?", to which the LLM responds definitively. Even when prompted for its confidence in its answer, the LLM claims that it is ``100\% confident." However, the LLM's answer fails to pass a fact-check test.
Knowing how much to trust an LLM-generated response is critical for users \cite{kim2024m}, helping inform the development of contingency strategies commensurate with the degree of uncertainty of the LLM in its response. For example, in applications such as robotics, an LLM-equipped robot could seek human guidance \cite{ren2023robots} or necessitate further review in the judicial practice \cite{delacroix2024augmenting}. Uncertainty quantification (UQ) methods for LLMs seek to address this challenge by providing users with an estimate of an LLM's confidence in its response to a given prompt. Indeed, uncertainty quantification can be important in factuality analysis \cite{huang2023look}.

The rapid adoption of LLMs in many applications has contributed to the fast-pace development of UQ methods for LLMs to promote their safe integration into a wide range of applications. However, the huge volume of UQ methods for LLMs has made it particularly challenging to ascertain the research scope and guarantees provided by existing UQ methods, complicating the identification of useful UQ methods for practitioners seeking to leverage them in their application areas, as well as the identification of impactful future directions for research. We claim that this challenge arises from the lack of a taxonomy that unifies related existing methods and presents an organized view of existing work in this research area. 

Through this survey, we seek not only to enumerate existing work in UQ for LLMs, but also to provide a useful taxonomy of UQ methods for LLMs to aid understanding the state of the art in this research area. We reiterate that the introduction of an effective taxonomy for these methods can facilitate their adoption in wide-ranging applications, such as in factuality analysis, hallucination detection, and robotics. \hypertarget{link:taxonomy}{We categorize existing uncertainty quantification methods for LLMs into four main classes: (1) token-level uncertainty quantification methods; (2) self-verbalized uncertainty quantification methods; (3) semantic-similarity uncertainty quantification methods; and (4) mechanistic interpretability methods. These categories encompass uncertainty quantification of multi-claim, multi-sentence LLM responses.} We elaborate on each category in this survey, identifying the key features shared by methods within each category. Moreover, we identify open research challenges and provide directions for future research, hoping to inspire future effort in advancing the state of the art.

\subsection*{Comparison to other Surveys}
A number of surveys on hallucinations in LLMs exists, e.g., \cite{rawte2023survey, huang2023survey, tonmoy2024comprehensive, liu2024survey, bai2024hallucination}. These surveys discuss hallucinations in detail, introducing the notion of hallucinations \cite{rawte2023survey}, identifying its types and potential causes \cite{huang2023survey}, and presenting mitigation techniques \cite{tonmoy2024comprehensive}. However, these papers provide little to no discussion on uncertainty quantification methods for LLMs, as this research area lies outside the scope of these surveys. 
In contrast, only two surveys on uncertainty quantification methods for LLMs exist, to the best of our knowledge. The first survey \cite{geng2024survey} categorizes confidence estimation and calibration methods into two broad classes: methods for generation tasks and methods for classification tasks, defined by the application domain. The survey in \cite{geng2024survey} focuses more heavily on calibration methods, with a less extensive discussion on confidence estimation methods. In contrast, our paper provides an extensive survey of uncertainty quantification methods with a brief discussion on calibration of uncertainty estimates. For example, whereas \cite{geng2024survey} lacks a detailed discussion on the emerging field of mechanistic interpretability, our survey presents this field in detail, along with potential applications to uncertainty quantification. Moreover, our survey discusses a broad range of applications of uncertainty quantification methods for LLMs, e.g., embodied applications such as in robotics, beyond those discussed in \cite{geng2024survey}.
A concurrent survey \cite{huang2024survey} on uncertainty quantification of LLMs categorizes existing uncertainty quantification methods within more traditional classes, which do not consider the unique architecture and characteristics of LLMs. In contrast, our survey categorizes existing work within the lens of LLMs, considering the underlying transformer architecture of LLMs and the autoregressive token-based procedure utilized in language generation.

\subsection*{Organization}
In Section~\ref{sec:background}, we begin with a review of essential concepts that are necessary for understanding the salient components of uncertainty quantification of LLMs. We discuss the general notion of uncertainty and introduce the main categories of uncertainty quantification methods within the broader field of deep learning. Subsequently, we identify the relevant metrics utilized by a majority of uncertainty quantification methods for LLMs. In Sections~\ref{sec:token_level_uq}, \ref{sec:self_verbalized_uq}, \ref{sec:semantic_similarity_uq}, and \ref{sec:mechanistic_interpretability_uq}, we discuss the four main categories of uncertainty quantification methods for LLMs, highlighting the key ideas leveraged by the methods in each category. In Section~\ref{sec:calibration}, we provide a brief discussion of calibration techniques for uncertainty quantification, with applications to uncertainty quantification of LLMs. In Section~\ref{sec:dataset}, we summarize the existing datasets and benchmarks for uncertainty quantification of LLMs and present applications of uncertainty quantification methods for LLMs in Section~\ref{sec:applications}. We highlight open challenges in Section~\ref{sec:open_research_challenges} and suggest directions for future research. Lastly, we provide concluding remarks in Section~\ref{sec:conclusion}. \Cref{fig:outline} summarizes the organization of this survey, highlighting the key details presented therein.

\begingroup
\hypersetup{linkcolor=black}

\begin{center}
    {
    \begin{adjustbox}{width=0.8\textwidth}
    {
        \begin{tikzpicture}[node distance=2cm]
        
        \node[rotate=90] (level_0) [rect] {Uncertainty Quantification for LLMs};
        
        \node (level_1_t0) [rect_wrap, fill=row0_color!90, right of=level_0, xshift=10ex, yshift=46ex] {\hyperlink{link:taxonomy}{Taxonomy}};
        \node (level_1_t1) [rect_wrap, fill=row1_color!90, right of=level_0, xshift=10ex, yshift=13ex] {\hyperref[sec:dataset]{Datasets and Benchmarks}};
        \node (level_1_t2) [rect_wrap, fill=row2_color!90, right of=level_0, xshift=10ex, yshift=-12ex] {\hyperref[sec:applications]{Applications}};
        \node (level_1_t3) [rect_wrap, fill=row3_color!90, right of=level_0, xshift=10ex, yshift=-41ex] {\hyperref[sec:open_research_challenges]{Open Challenges and Future Directions}};
        
        \draw [arrow] (level_0.south) -- ++(2.5ex, 0ex) |- (level_1_t0.west);
        \draw [arrow] (level_0.south) -- ++(2.5ex, 0ex) |- (level_1_t1.west);
        \draw [arrow] (level_0.south) -- ++(2.5ex, 0ex) |- (level_1_t2.west);
        \draw [arrow] (level_0.south) -- ++(2.5ex, 0ex) |- (level_1_t3.west);
        
        \node (level_2_t0_0) [rect_large, fill=row0_color!50, right of=level_1_t0, xshift=25ex, yshift=12ex] {\hyperref[sec:token_level_uq]{Token-Level UQ}};
        \node (level_2_t0_1) [rect_large, fill=row0_color!50, right of=level_1_t0, xshift=25ex, yshift=4ex] {\hyperref[sec:self_verbalized_uq]{Self-Verbalized UQ}};
        \node (level_2_t0_2) [rect_large, fill=row0_color!50, right of=level_1_t0, xshift=25ex, yshift=-4ex] {\hyperref[sec:semantic_similarity_uq]{Semantic-Similarity UQ}};
        \node (level_2_t0_3) [rect_large, fill=row0_color!50, right of=level_1_t0, xshift=25ex, yshift=-12ex] {\hyperref[sec:mechanistic_interpretability_uq]{Mechanistic Interpretability}};
        
        \draw [arrow] (level_1_t0.east) -- ++(2.5ex, 0ex) |- (level_2_t0_0.west);
        \draw [arrow] (level_1_t0.east) -- ++(2.5ex, 0ex) |- (level_2_t0_1.west);
        \draw [arrow] (level_1_t0.east) -- ++(2.5ex, 0ex) |- (level_2_t0_2.west);
        \draw [arrow] (level_1_t0.east) -- ++(2.5ex, 0ex) |- (level_2_t0_3.west);
        
        \node (level_2_t1_0) [rect_large, fill=row1_color!50, right of=level_1_t1, xshift=25ex, yshift=12ex] {\hyperlink{link:read_comp_benchmark}{Reading Comprehension}};
        \node (level_2_t1_1) [rect_large, fill=row1_color!50, right of=level_1_t1, xshift=25ex, yshift=4ex] {\hyperlink{link:math_benchmark}{Mathematics}};
        \node (level_2_t1_2) [rect_large, fill=row1_color!50, right of=level_1_t1, xshift=25ex, yshift=-4ex] {\hyperlink{link:multi_hop_reasoning}{Multi-Hop Reasoning}};
        \node (level_2_t1_3) [rect_large, fill=row1_color!50, right of=level_1_t1, xshift=25ex, yshift=-12ex] {\hyperlink{link:factuality_analysis}{Factuality Analysis}};
        
        \draw [arrow] (level_1_t1.east) -- ++(2.5ex, 0ex) |- (level_2_t1_0.west);
        \draw [arrow] (level_1_t1.east) -- ++(2.5ex, 0ex) |- (level_2_t1_1.west);
        \draw [arrow] (level_1_t1.east) -- ++(2.5ex, 0ex) |- (level_2_t1_2.west);
        \draw [arrow] (level_1_t1.east) -- ++(2.5ex, 0ex) |- (level_2_t1_3.west);
        
        \node (level_2_t2_1) [rect_large, fill=row2_color!50, right of=level_1_t2, xshift=25ex, yshift=4ex] {\hyperref[ssec:chatbot_applications]{Chatbot and Textual}};
        \node (level_2_t2_0) [rect_large, fill=row2_color!50, right of=level_1_t2, xshift=25ex, yshift=-4ex] {\hyperref[ssec:robotics_applications]{Robotics}};
        
        \draw [arrow] (level_1_t2.east) -- ++(2.5ex, 0ex) |- (level_2_t2_0.west);
        \draw [arrow] (level_1_t2.east) -- ++(2.5ex, 0ex) |- (level_2_t2_1.west);
        
        \node (level_2_t3_0) [rect_larger, fill=row3_color!50, right of=level_1_t3, xshift=32.5ex, yshift=16ex] {\hyperref[ssec:challenges_consistency]{Consistency and Factuality}};
        \node (level_2_t3_1) [rect_larger, fill=row3_color!50, right of=level_1_t3, xshift=32.5ex, yshift=8ex] {\hyperref[ssec:challenges_entropy]{Entropy and Factuality}};
        \node (level_2_t3_2) [rect_larger, fill=row3_color!50, right of=level_1_t3, xshift=32.5ex, yshift=0ex] {\hyperref[ssec:challenges_interactive_llm]{Multi-Episode UQ for Interactive Agents}};
        \node (level_2_t3_3) [rect_larger, fill=row3_color!50, right of=level_1_t3, xshift=32.5ex, yshift=-8ex] {\hyperref[ssec:challenges_mechanical interpretability]{Mechanistic Interpretability and UQ}};
        \node (level_2_t3_4) [rect_larger, fill=row3_color!50, right of=level_1_t3, xshift=32.5ex, yshift=-16ex] {\hyperref[ssec:challenges_datasets_benchmark]{Datasets and Benchmarks}};
        
        \draw [arrow] (level_1_t3.east) -- ++(2.5ex, 0ex) |- (level_2_t3_0.west);
        \draw [arrow] (level_1_t3.east) -- ++(2.5ex, 0ex) |- (level_2_t3_1.west);
        \draw [arrow] (level_1_t3.east) -- ++(2.5ex, 0ex) |- (level_2_t3_2.west);
        \draw [arrow] (level_1_t3.east) -- ++(2.5ex, 0ex) |- (level_2_t3_3.west);
        \draw [arrow] (level_1_t3.east) -- ++(2.5ex, 0ex) |- (level_2_t3_4.west);
        
        \node (level_3_t0_0) [rect_medium, fill=row0_color!20, align=left, right of=level_2_t0_0, xshift=27ex, yshift=0ex] { \hspace{0.2ex} \cite{xiao2021hallucination, kadavath2022language, bakman2024mars, ling2024uncertainty, vazhentsev2024unconditional, fadeeva2024fact, ren2023a}}; %
        \node (level_3_t0_1) [rect_medium, fill=row0_color!20, align=left, right of=level_2_t0_1, xshift=27ex, yshift=0ex] {\hspace{0.2ex}  \cite{mielke2022reducing, lin2022teaching, stengel2024lacie, yang2024can, xu2024sayself, tao2024trust, band2024linguistic}};
        \node (level_3_t0_2) [rect_medium, fill=row0_color!20, align=left, right of=level_2_t0_2, xshift=27ex, yshift=0ex] {\hspace{0.2ex}  \cite{kuhn2023semantic, chen2023quantifying, lin2023generating, kossen2024semantic, wang2024clue, qiu2024semantic, ao2024css}};
        \node (level_3_t0_3) [rect_medium, fill=row0_color!20, align=left, right of=level_2_t0_3, xshift=27ex, yshift=0ex] { \hspace{0.2ex} \cite{ahdritz2024distinguishing}};
        
        \draw [arrow] (level_2_t0_0.east) -- (level_3_t0_0.west);
        \draw [arrow] (level_2_t0_1.east) -- (level_3_t0_1.west);
        \draw [arrow] (level_2_t0_2.east) -- (level_3_t0_2.west);
        \draw [arrow] (level_2_t0_3.east) -- (level_3_t0_3.west);
        
        \node (level_3_t1_0) [rect_medium, fill=row1_color!20, align=left, right of=level_2_t1_0, xshift=27ex, yshift=0ex] {\hspace{0.2ex}  \cite{joshi2017triviaqa, reddy2019coqa, lebret2016generating}};
        \node (level_3_t1_1) [rect_medium, fill=row1_color!20, align=left, right of=level_2_t1_1, xshift=27ex, yshift=0ex] {\hspace{0.2ex}  \cite{lin2022teaching}};
        \node (level_3_t1_2) [rect_medium, fill=row1_color!20, align=left, right of=level_2_t1_2, xshift=27ex, yshift=0ex] {\hspace{0.2ex}  \cite{yang2018hotpotqa, geva2021did}};
        \node (level_3_t1_3) [rect_medium, fill=row1_color!20, align=left, right of=level_2_t1_3, xshift=27ex, yshift=0ex] {\hspace{0.2ex}  \cite{lin2021truthfulqa, li2023halueval, thorne2018fever}};
        
        \draw [arrow] (level_2_t1_0.east) -- (level_3_t1_0.west);
        \draw [arrow] (level_2_t1_1.east) -- (level_3_t1_1.west);
        \draw [arrow] (level_2_t1_2.east) -- (level_3_t1_2.west);
        \draw [arrow] (level_2_t1_3.east) -- (level_3_t1_3.west);
        
        \node (level_3_t2_0) [rect_medium, fill=row2_color!20, align=left, right of=level_2_t2_0, xshift=27ex, yshift=0ex] {\hspace{0.2ex}  \cite{tsai2024efficient, ren2023robots, Wang2023ConformalTL, liang2024introspective, mullen2024towards, wang2024safe, zheng2024evaluating}};
        \node (level_3_t2_1) [rect_medium, fill=row2_color!20, align=left, right of=level_2_t2_1, xshift=27ex, yshift=0ex] {\hspace{0.2ex} \cite{zhang2023enhancing, yadkori2024believe, mohri2024language, pacchiardi2023catch, tai2024examination, kolagar2024aligning, steindl2024linguistic}};
        
        \draw [arrow] (level_2_t2_0.east) -- (level_3_t2_0.west);
        \draw [arrow] (level_2_t2_1.east) -- (level_3_t2_1.west);

        \end{tikzpicture}
        }
    \end{adjustbox}
    }
    
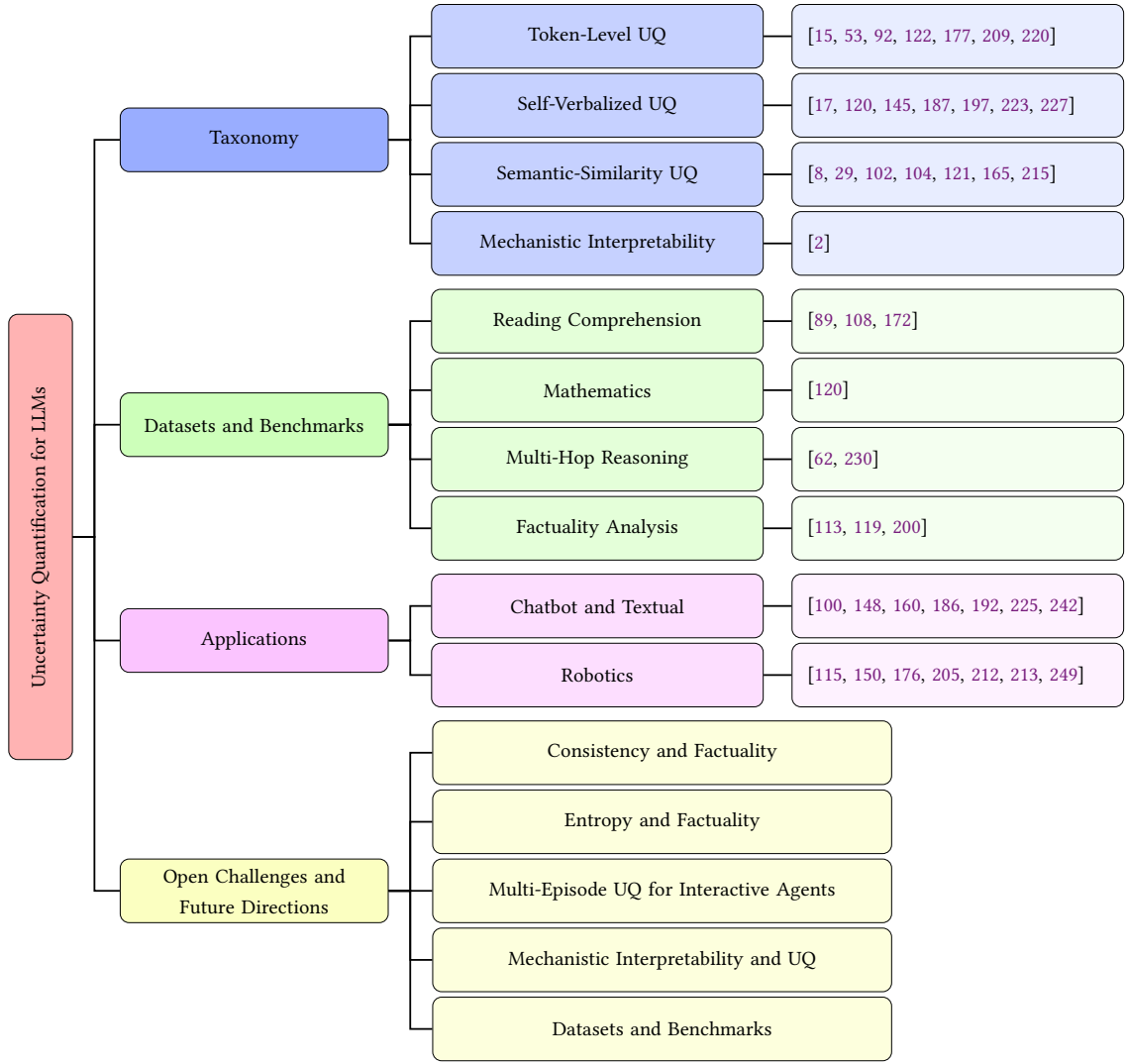
\captionof{figure}{The overview of this survey, including a taxonomy of uncertainty quantification methods for LLMs, relevant datasets and benchmarks, applications, and open challenges and directions for future research.}
    \label{fig:outline}
\end{center}
\endgroup

\section{Background}
\label{sec:background}
We review fundamental concepts that are crucial to understanding uncertainty quantification of LLMs. We assume basic familiarity with deep learning and build upon this foundation to introduce more specific concepts, describing the notion of uncertainty, the inner workings of LLMs, and the development of metrics and probes to illuminate the uncertainty of LLMs in their response to a user's prompt.

\subsection{Uncertainty}
Uncertainty is a widely-known, yet vaguely-defined concept. For example, people generally associate uncertainty with doubt or a lack of understanding, knowledge, or control, but cannot generally provide a precise definition, especially a mathematical one. This general ambiguity applies to the field of LLMs \cite{keeling2024attribution}. For example, a subset of the LLM research field considers the uncertainty of a model to be distinct from its level of confidence in a response generated by the model \cite{lin2023generating}, stating that confidence scores are associated with a prompt (input) and a prediction by the model, whereas uncertainty is independent of the model's prediction.
However, a large subset of the field considers uncertainty and the lack of confidence to be mostly-related, generally-interchangeable concepts. In this section, for simplicity, we consider uncertainty and confidence to be mostly interchangeable.

\begin{figure}[th]
    \centering
    \begin{minipage}[b][][b]{.48\textwidth}
        \centering
        \includegraphics[width=\columnwidth]{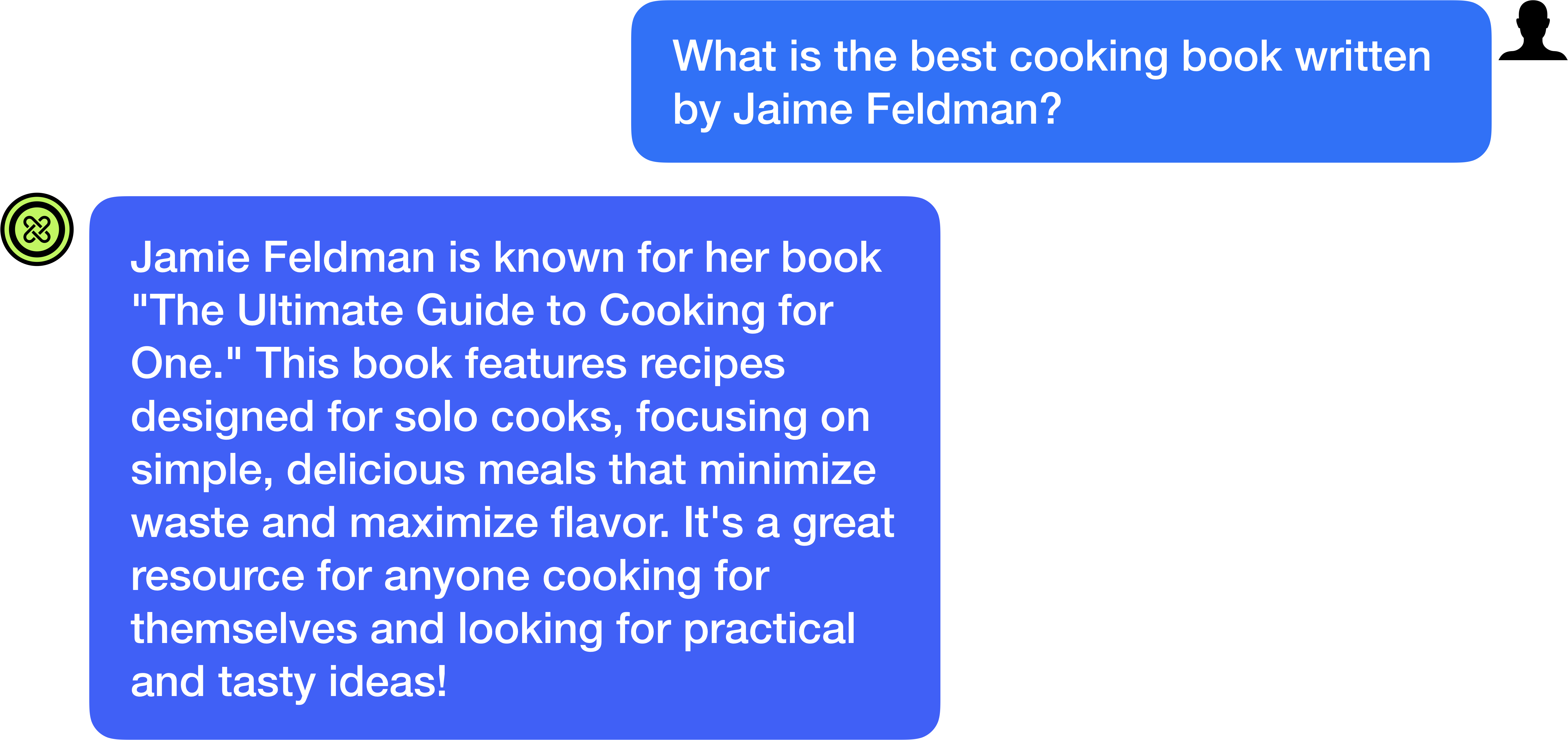}
        \caption{Hallucination in  LLMs: When asked for information about a possibly fictional person, LLMs tend to fabricate a response that sounds coherent but is entirely false.}
        \Description[Hallucination in  LLMs: When asked for information about a possibly fictional person, LLMs tend to fabricate a response that sounds coherent but is entirely false.]{Hallucination in  LLMs: When asked for information about a possibly fictional person, LLMs tend to fabricate a response that sounds coherent but is entirely false.}
        \label{fig:hallucination_cooking_book}
    \end{minipage}%
    \hfill %
    \begin{minipage}[b][][b]{.48\textwidth}
        \centering
        \includegraphics[width=\columnwidth]{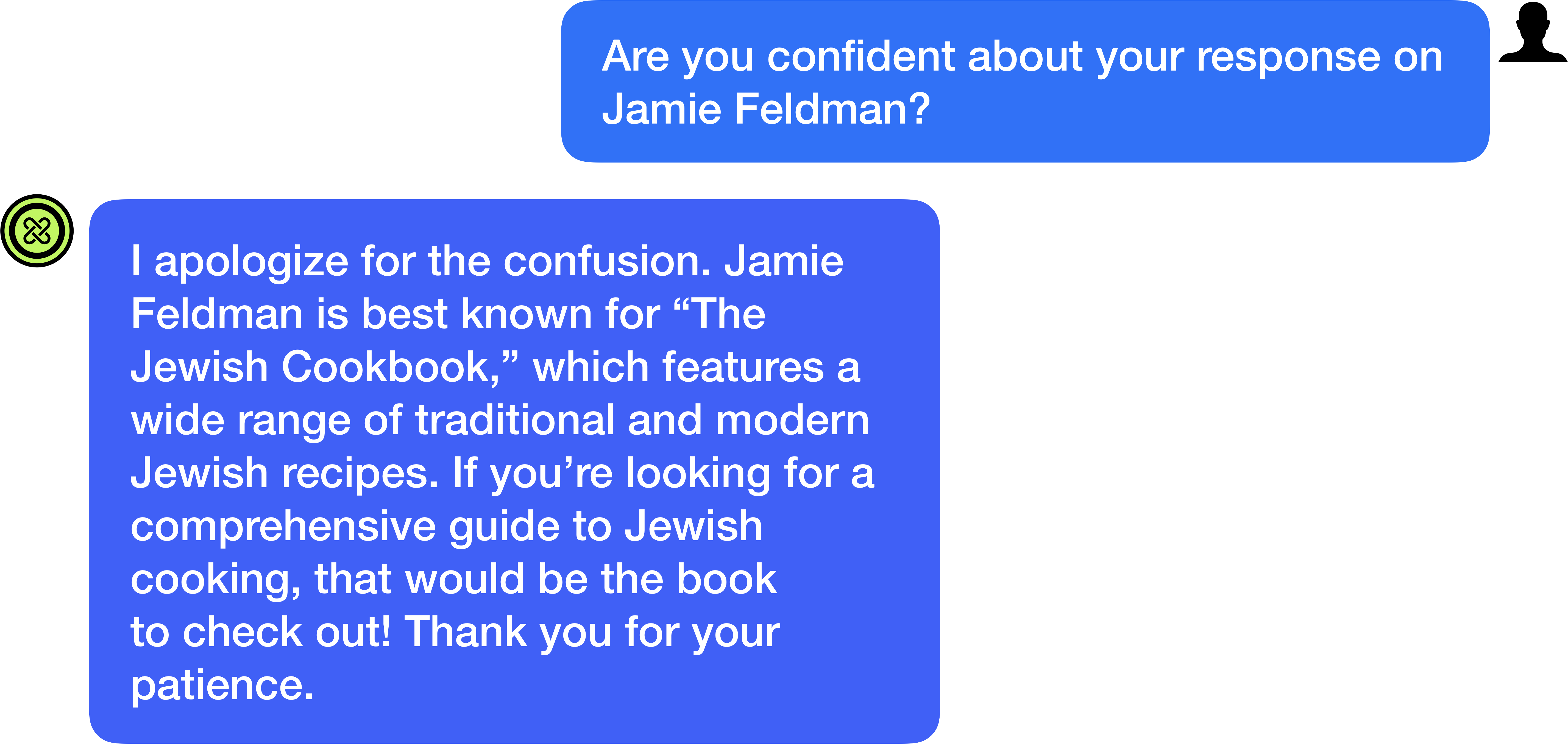}
        \caption{Hallucination in  LLMs: When asked about its confidence, the LLM apologizes before hallucinating another response. The Jewish Cookbook is authored by \emph{Leah Koenig}, not Jaime Feldman.}
        \Description[Hallucination in  LLMs: When asked about its confidence, the LLM apologizes before hallucinating another response. The Jewish Cookbook is authored by \emph{Leah Koenig}, not Jaime Feldman.]{Hallucination in  LLMs: When asked about its confidence, the LLM apologizes before hallucinating another response. The Jewish Cookbook is authored by \emph{Leah Koenig}, not Jaime Feldman.}
        \label{fig:hallucination_cooking_book_v1}
    \end{minipage}
\end{figure}

When prompted, LLMs tend to hallucinate when uncertainty about the correct answer exists, e.g., when a lack of understanding or a lack of knowledge exists (see \Cref{fig:hallucination_cooking_book,fig:hallucination_cooking_book_v1}). In \Cref{fig:hallucination_cooking_book,fig:hallucination_cooking_book_v1},
we ask GPT-4o mini to name the best cooking book written by a (likely) fictional person Jamie Feldman. GPT-4o mini provides a confident response: ``The Ultimate Guide to Cooking for One." However, based on an internet search, this cookbook does not exist (although many similar ones do). Moreover, when prompted about its confidence, GPT-4o mini apologizes before providing yet another confident, but factually-incorrect response: ``The Jewish Cookbook." This book is authored by \emph{Leah Koenig}, not Jamie Feldman. Uncertainty quantification (UQ) methods aim to provide a more rigorous estimate of the model's confidence in its response, e.g., from the entropy of the distribution from which the tokens are sampled. Before discussing UQ techniques for LLMs, we identify the types of uncertainty and the methods suitable for characterizing uncertainty in deep-learned models, more broadly.

\subsection{Types of Uncertainty} 
Uncertainty can be broadly categorized into two classes, namely: \emph{aleatoric} uncertainty and \emph{epistemic} uncertainty. When considered collectively, the resulting uncertainty is referred to as \emph{predictive} uncertainty, without a distinction between the two components.

\subsubsection{Aleatoric Uncertainty}
Aleatoric uncertainty encompasses the lack of definiteness of the outcome of an event due to the inherent randomness in the process which determines the outcome of the event. For example, a model cannot predict with certainty the outcome of an unbiased coin toss due to the random effects in the coin toss, regardless of the complexity of the model or the size of the training dataset used in training the model. This irreducible uncertainty is referred to as aleatoric uncertainty. 
For example, in the case of LLMs, aleatoric uncertainty can arise when there is inherent randomness in the ground-truth response, e.g., when prompted with ``What will the temperature be tomorrow?", the uncertainty associated with the LLM's output can be characterized as aleatoric uncertainty, which is entirely due to the random effects associated with daily weather conditions. In essence, daily weather conditions cannot be predicted with absolute certainty, irrespective of the amount of training data available.

\subsubsection{Epistemic Uncertainty}
In contrast to aleatoric uncertainty, epistemic uncertainty characterizes the doubt associated with a certain outcome (prediction) due to a lack of knowledge or ``ignorance" by a model, often due to limited training data. 
For example, when prompted to provide the digit in the $7$th decimal place in the square-root of $2$, GPT-4o~mini responds with the answer $6$. However, this answer is wrong: the digit in the $7$th decimal place is $5$. The uncertainty in the LLM's output can be characterized as epistemic uncertainty, which can be eliminated by training the LLM on more data specific to this prompt.
In other words, epistemic uncertainty describes reducible uncertainty, i.e., epistemic uncertainty
should reduce when there is more knowledge about the state on which the decision is being made, e.g., via choosing the right model to use for learning, using more training data, or by incorporating any prior knowledge.
The uncertainty associated with the response in \Cref{fig:hallucination_cooking_book} is entirely epistemic and stems from missing training data. If we train the LLM on more data, including the fact that Jamie Feldman did not write a cookbook, we can eliminate the uncertainty associated with the model's response. 
Before concluding, we note that prior work has explored decomposing predictive uncertainty into epistemic and aleatoric components~\cite{hou2023decomposing}.

\subsection{Uncertainty Quantification in Deep Learning}

Broadly, uncertainty quantification for deep learning lies along a spectrum between two extremes: \emph{training-based} and \emph{training-free}  methods, illustrated in Figure~\ref{fig:UQinDeepLearning}. 
Whereas training-based methods assume partial or complete visibility and access to the internal structure of the neural network, modifying it to probe its uncertainty, training-free methods use auxiliary models or additional data to quantify the uncertainty of the model post-hoc.

\begin{center}
{
\begin{adjustbox}{width=0.8\textwidth}
{
\begin{tikzpicture}

\definecolor{yellowhighlight}{RGB}{255, 235, 156}
\definecolor{greenhighlight}{RGB}{198, 239, 206}
\definecolor{bluehighlight}{RGB}{221, 235, 247}

\draw[thick,<->] (0,3.5) -- (15,3.5);
\node[anchor=south] at (1.3,3.5) {\textbf{Training-based}};
\node[anchor=south] at (13.85,3.5) {\textbf{Training-free}};

\draw[thick] (0,0) rectangle (7.15,3.1);
\draw[thick]  (0.15,0.15) rectangle (2.35,2.5);
\draw[thick]  (2.5,0.15) rectangle (7,2.5);
\draw[thick]  (2.65,0.3) rectangle (6.85,0.8);
\draw[thick]  (2.65,0.95) rectangle (6.85,1.95);

\node[anchor=north west,rounded corners] at (0.1,3.1) {\textbf{BNNs}~\cite{jospin2022hands}};
\node[anchor=north west,rounded corners] at (0.15,2.49) {MCMC~\cite{hastings1970monte}};
\node[anchor=north west,rounded corners] at (2.5,2.5) {{Variational Inference}~\cite{posch2019variationalinferencemeasuremodel}};
\node[anchor=north west,rounded corners] at (2.7,0.85) {MC-Dropout~\cite{gal2016dropout, gal2017concrete}};
\node[anchor=north west,rounded corners] at (2.48,2.10) {\begin{tabular}{l}
     Deep Ensemble  \\[-0.1cm]
     \cite{lakshminarayanan2017simple, guo2018margin, cavalcanti2016combining, martinez2008analysis, buciluǎ2006model, hinton2015distilling}
\end{tabular}};

\draw[thick] (7.4,0) rectangle (9.2,3.1);
\node[anchor=north west,rounded corners] at (7.22,3.1) {\begin{tabular}{l}
    \textbf{ENNs}  \\
    \cite{osband2023epistemic, wang2024epistemic} 
\end{tabular}};

\draw[thick] (9.45,2) rectangle (15,3.1);
\draw[thick] (9.45,1) rectangle (15,1.9);
\draw[thick] (9.45,0) rectangle (15,0.9);

\node[anchor=north west,rounded corners] at (9.34,3.215) {\begin{tabular}{l}
    Test-time Data Augmentation \\[-0.1cm]
    \cite{lee2020gradients, ayhan2018test, wu2024posterior, bahat2020classification}
\end{tabular}};
\node[anchor=north west,rounded corners] at (9.5,1.8) {Dropout Injection~\cite{loquercio2020general, ledda2023dropout}};
\node[anchor=north west,rounded corners] at (9.5,0.85) {Gradient-based~\cite{lee2020gradients, huang2021importance, igoe2022useful}};

\end{tikzpicture}
}
\end{adjustbox}
}

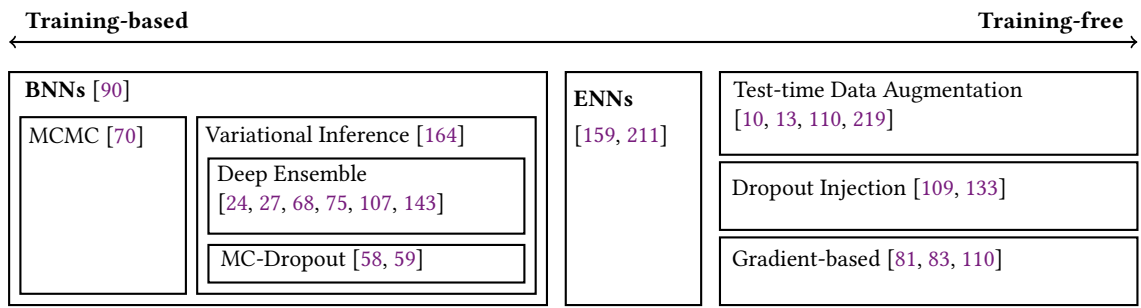
\captionof{figure}{Uncertainty quantification methods in deep learning span the spectrum from training-based methods to training-free methods.}
\label{fig:UQinDeepLearning}
\end{center}

\subsubsection{Training-Based Methods}
Training-based uncertainty quantification methods span Bayesian Neural Networks, Monte Carlo Dropout methods, and Deep Ensembles, which we review in the subsequent discussion. 
Instead of training a set of parameters to predict a single outcome, a Bayesian neural network (BNN) \cite{jospin2022hands} learns a distribution over the model's weights $\theta$. 
Specifically, a BNN learns a distribution over the parameters,  $p(\theta|D)$, 
with dataset $D$,
with its prediction consisting of two parts: a maximum a posteriori estimation component $\hat y$, and the uncertainty associated with it, defined by the covariance of the prediction $\Sigma_{\hat y|x, D}$.

Despite being statistically principled, the prohibitive computational costs associated with BNNs prevent them from being directly employed. In order to train BNNs, a variety of methods have been proposed, among which the most popular ones are Markov Chain Monte-Carlo (MCMC)~\cite{hastings1970monte} and variational inference~\cite{posch2019variationalinferencemeasuremodel}. The former samples from the exact posterior distribution, while the latter learns to approximate the posterior with a variational distribution,  $q_\varphi$. Due to the relaxed requirement of access to large amounts of samples, the variational inference method has been more widely used, with Monte-Carlo dropout~\cite{gal2016dropout, gal2017concrete} and Deep ensemble~\cite{lakshminarayanan2017simple} being representative methods. More recently, epistemic neural networks (ENNs)~\cite{osband2023epistemic, wang2024epistemic} have been introduced to reduce the computational challenges associated with BNNs. To make ensemble methods more efficient, e.g., in out-of-distribution detection \cite{vyas2018out}, pruning methods~\cite{guo2018margin, cavalcanti2016combining, martinez2008analysis}, which reduce redundancy among ensemble members, and distillation methods~\cite{buciluǎ2006model, hinton2015distilling}, which reduce the number of networks to one, teaching it to represent the knowledge of a group of networks, have been introduced.
While these methods are easy to implement and require much less computation compared to regular BNNs or MCMC, they do suffer from being an approximation of the true posterior distribution. In fact, the model's uncertainty predictions could be worse when data augmentation, ensembling, and post-processing calibration are used together~\cite{rahaman2021uncertainty}.

\subsubsection{Training-Free Methods}

Training-free methods for estimating uncertainty have become popular due to their ease of implementation. Since neither the network architecture nor the training process need to be revised, training-free methods work well with large-scale foundation models that are costly to train or fine-tune. 
In~\cite{ayhan2018test, lee2020gradients, wu2024posterior, bahat2020classification}, the authors perform data augmentation at test time to generate a predictive distribution, quantifying the model's uncertainty.
Similarly, dropout injection~\cite{loquercio2020general, ledda2023dropout} extends MC-dropout to the training-free domain by only performing dropout at inference time to estimate epistemic uncertainty. In~\cite{mi2022training}, the authors estimate uncertainty for regression using similar perturbation techniques.
Lastly, gradient-based uncertainty quantification methods~\cite{lee2020gradients} generate gradients at test-time, which provide an signal for epistemic uncertainty and for OOD detection in~\cite{huang2021importance, igoe2022useful},  by constructing confounding labels.

\subsection{Uncertainty Quantification for LLMs} 
\label{sec:background_uq_llm}
The introduction of the transformer \cite{vaswani2017attention} for sequence-to-sequence machine translation tasks spurred the development of large language models. However, as noted in the preceding discussion, LLMs have gained some notoriety for their tendency to hallucinate when uncertain about a response to a specified prompt. Here, we review the general architecture of LLMs and provide some motivation for the development of LLM-specific metrics for quantifying uncertainty.

\subsubsection{LLM Architecture}
LLMs use the transformer architecture to provide free-form responses to input prompts specified in natural language.
The transformer architecture consists of 
an encoder, which processes the input to the model, and a decoder, which generates the model's outputs auto-regressively, where the previous outputs of the model are passed into the model to generate the future outputs. 
Given an input prompt, the words (elements) of the prompt are tokenized (i.e., the sentences/phrases in natural-language are decomposed into simple units referred to as tokens) and transformed to input embeddings using a learned model. The encoder takes in the input embeddings augmented with positional encodings to incorporate positional context and generates a sequence of latent embeddings, which serves as an input to the decoder, using a stack of $N$ multi-head attention sub-blocks and fully-connected feedforward networks.  
The decoder takes in the embeddings associated with the  previous outputs of the decoder, preceded by a \emph{start} token, and computes an output embedding using a similar stack of multi-head attention heads and feedforward networks as the encoder. The resulting output embeddings are passed into a linear layer prior to a softmax output layer, which converts the decoder embeddings to a probability distribution over the tokens in the dictionary of the model. In subsequent discussion, we denote the probability of the $j$'th token in the $i$'th sentence of an LLM's output as $p_{ij}$. The output token is selected from this probability distribution: e.g., by greedily taking the token associated with the maximum probability mass. The resulting output is passed into the decoder for auto-regressive generation of text.

\begin{figure}[th]
    \centering
    \includegraphics[width=\textwidth]{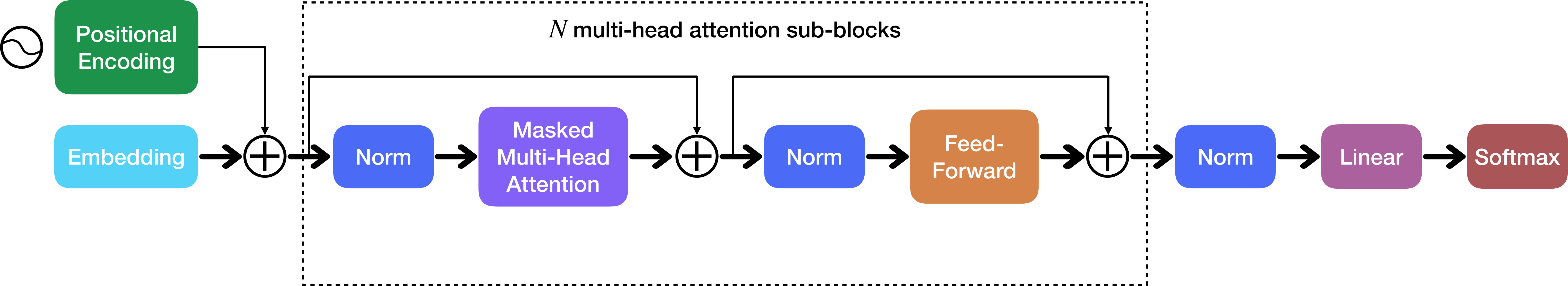}
    \caption{Many state-of-the-art LLMs are decoder-only transformers, with $N$ multi-head attention sub-blocks, for auto-regressive output generation.}
    \Description[Many state-of-the-art LLMs are decoder-only transformers, with $N$ multi-head attention sub-blocks, for auto-regressive output generation.]{Many state-of-the-art LLMs are decoder-only transformers, with $N$ multi-head attention sub-blocks, for auto-regressive output generation.}
    \label{fig:transformer_architecture}
\end{figure}

While early LLM models utilized encoder-only or encoder-decoder transformer architectures, state-of-the-art LLMs now generally utilize a decoder-only architecture. For example, the GPT family of models, such as GPT-4 \cite{achiam2023gpt}, and the Llama family of models, such as Llama 3 \cite{dubey2024llama}, are decoder-only transformers. In \Cref{fig:transformer_architecture}, we show a decoder-only transformer model. These state-of-the-art models leverage advances in transformers to improve computational efficiency, given the huge size of these models: Llama~$3$ has $8$B parameters for the small variant and $70$B parameters for the large variant, while GPT-$4$ is rumored to have over one trillion parameters. Llama~$3$ uses rotary positional embeddings (RoPE) \cite{su2024roformer} instead of absolute positional embeddings, which have been shown to be more effective than alternative embedding schemes.
For a more detailed review of LLMs, we refer readers to \cite{minaee2024large}.
Before presenting the metrics utilized by UQ methods for LLMs, we discuss natural-language inference, which is an important component of many UQ methods for LLMs.

\subsubsection{Natural-Language Inference}
\label{sec:nli_background}
\begin{figure}[th]
    \centering
    \includegraphics[width=0.73\columnwidth]{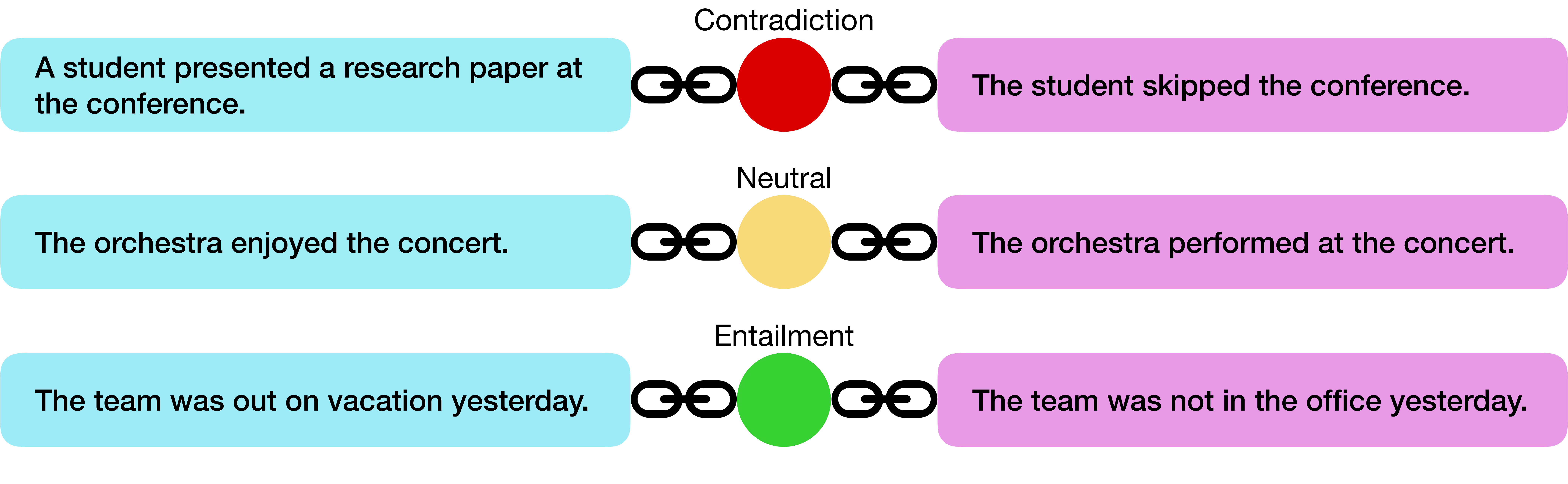}
    \caption{Natural-language inference models characterize the relationship between a pair of texts, namely: a premise and a hypothesis. The possible relations include: (1) an entailment where the hypothesis can be inferred from the premise; (2) a contradiction, where the hypothesis is more likely false given the premise; and (3) a neutral relation, where the veracity of the hypothesis cannot be determined from the premise.}
    \Description[Natural-language inference models characterize the relationship between a pair of texts, namely: a premise and a hypothesis. The possible relations include: (1) an entailment where the hypothesis can be inferred from the premise; (2) a contradiction, where the hypothesis is more likely false given the premise; and (3) a neutral relation, where the veracity of the hypothesis cannot be determined from the premise.]{Natural-language inference models characterize the relationship between a pair of texts, namely: a premise and a hypothesis. The possible relations include: (1) an entailment where the hypothesis can be inferred from the premise; (2) a contradiction, where the hypothesis is more likely false given the premise; and (3) a neutral relation, where the veracity of the hypothesis cannot be determined from the premise.}
    \label{fig:nli_labels}
\end{figure}

Natural-language inference (NLI) refers to the task of characterizing the relationship between two text fragments, where one text fragment represents a premise (i.e., a statement that is believed to be true) while the other fragment represents a hypothesis (i.e., a statement whose veracity we seek to evaluate based on the premise) \cite{williams2017broad, dagan2005pascal, fyodorov2000natural}. Given a premise and a hypothesis, we can classify the relation between the text pair as: an \emph{entailment}, if one can infer that the hypothesis is most likely true given the premise; a \emph{contradiction}, if one can infer that the hypothesis is most likely false given the premise; or a \emph{neutral} label, if one cannot infer the truthfulness of the hypothesis from the premise \cite{maccartney2008modeling, condoravdi2003entailment, monz2001light}. In \Cref{fig:nli_labels}, we provide some examples of text pairs that exhibit entailment, contradiction, or neutrality. In the first example, the premise indicates that the student \emph{presented} a research paper at a conference (i.e., the student \emph{did not} skip the conference), hence, the contradiction. In the second example, the premise indicates that the orchestra enjoyed the concert, but does \emph{not} state whether the orchestra performed at the concert (or just observed the event), hence the neutral label. In the third example, we can infer that the hypothesis is true, since the premise indicates that the team was on vacation, hence, \emph{not} in the office.

NLI methods play an important role in uncertainty quantification of LLMs. Many UQ methods for LLMs rely on characterization of the semantic relationship between multiple realizations of the LLM's responses to a given input prompt to determine the confidence of the model. Many of these methods rely on learned models for natural-language inference, such as BERT \cite{devlin2018bert}, which utilizes a 
transformer-based architecture to learn useful language representations that are crucial in natural-language tasks such as question answering and natural-language inference. Unlike many standard language models, e.g., Generative Pre-trained Transformer (GPT) \cite{Radford2018ImprovingLU}, which impose a unidirectionality constraint where every token can only attend to previous tokens, BERT employs a bidirectional approach where each token can attend to any token regardless of its relative position, using a masked language model, potentially enabling the model to capture broader context, especially in sentence-level tasks. In \cite{liu2019roberta}, the authors demonstrate that the performance of BERT is limited by inadequate pre-training and propose an improved model, named RoBERTa \cite{liu2019roberta}, which retains the same architecture as BERT but is trained for longer with larger mini-batches of data with longer sequences. DeBERTa \cite{he2020deberta} further improves upon the performance of RoBERTa by introducing a disentangled attention mechanism and an enhanced mask decoder.

\subsubsection{Metrics for Uncertainty Quantification for LLMs.}
\label{ssec:uq_metrics}
Uncertainty quantification in the LLM community has largely eschewed traditional UQ methods for learned models due to the notable computation cost of running inference on LLMs \cite{balabanov2024uncertainty}, although, a few UQ methods for LLMs utilize deep ensembles, e.g., \cite{wang2023lora, balabanov2024uncertainty, zhang2024luq, arteaga2024hallucination}, generally based on low-rank adaptation (LoRA) \cite{hu2021lora}. Consequently, many UQ methods in this space have introduced less computationally intensive approximate quantification methods that directly harness the unique architecture of LLM models to assess the uncertainty of these models. In some cases, these methods retain the high-level idea of ensemble methods, quantifying the uncertainty of the model on a given prompt using the outputs of a set of individual models or a collection of outputs from the same model, with a temperature parameter less than one to promote greater stochasticity in the tokens generated by the model. UQ methods for LLMs can be broadly categorized into white-box models and black-box models \cite{liu2024uncertainty, vashurin2024benchmarking}, illustrated in
\Cref{fig:white_box_uq} and \Cref{fig:black_box_uq}, respectively. 

\paragraph{White-Box UQ Methods}
\phantomsection
\label{par:white_box_uq_llm}
White-box UQ models assume that the underlying architecture of the model is partially or completely visible and accessible---hence the term \emph{white-box}---taking advantage of access to the intermediate outputs of the underlying models, such as the probability distribution over the generated tokens or outputs at the inner layers of the model, to assess the uncertainty of the model \cite{kuhn2023semantic, azaria2023internal, fadeeva2024fact}. 
We provide some metrics utilized by white-box UQ methods for LLMs, where $p_{ij}$ denotes the conditional probability of token $j$ (conditioned on all preceding tokens) in sentence $i$:

\begin{figure}[th]
    \centering
    \begin{minipage}[b][][b]{.48\textwidth}
        \centering
        \includegraphics[width=\columnwidth]{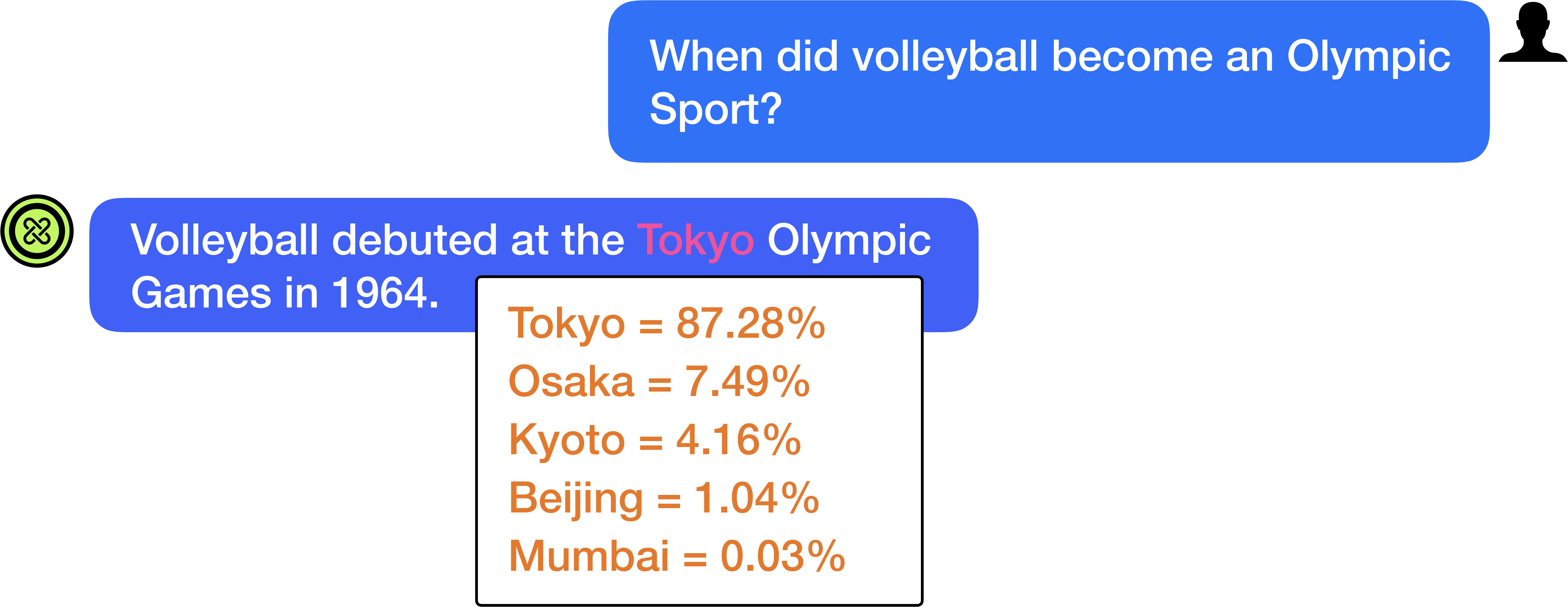}
        
        \caption{White-box uncertainty quantification methods utilize an LLM's internal information, e.g., the model's probabilities for the token associated with each output.}
        \Description[White-box uncertainty quantification methods utilize an LLM's internal information, e.g., the model's probabilities for the token associated with each output.]{White-box uncertainty quantification methods utilize an LLM's internal information, e.g., the model's probabilities for the token associated with each output.}
        \label{fig:white_box_uq}
    \end{minipage}%
    \hfill %
    \begin{minipage}[b][][b]{.48\textwidth}
        \centering
        \includegraphics[width=\columnwidth]{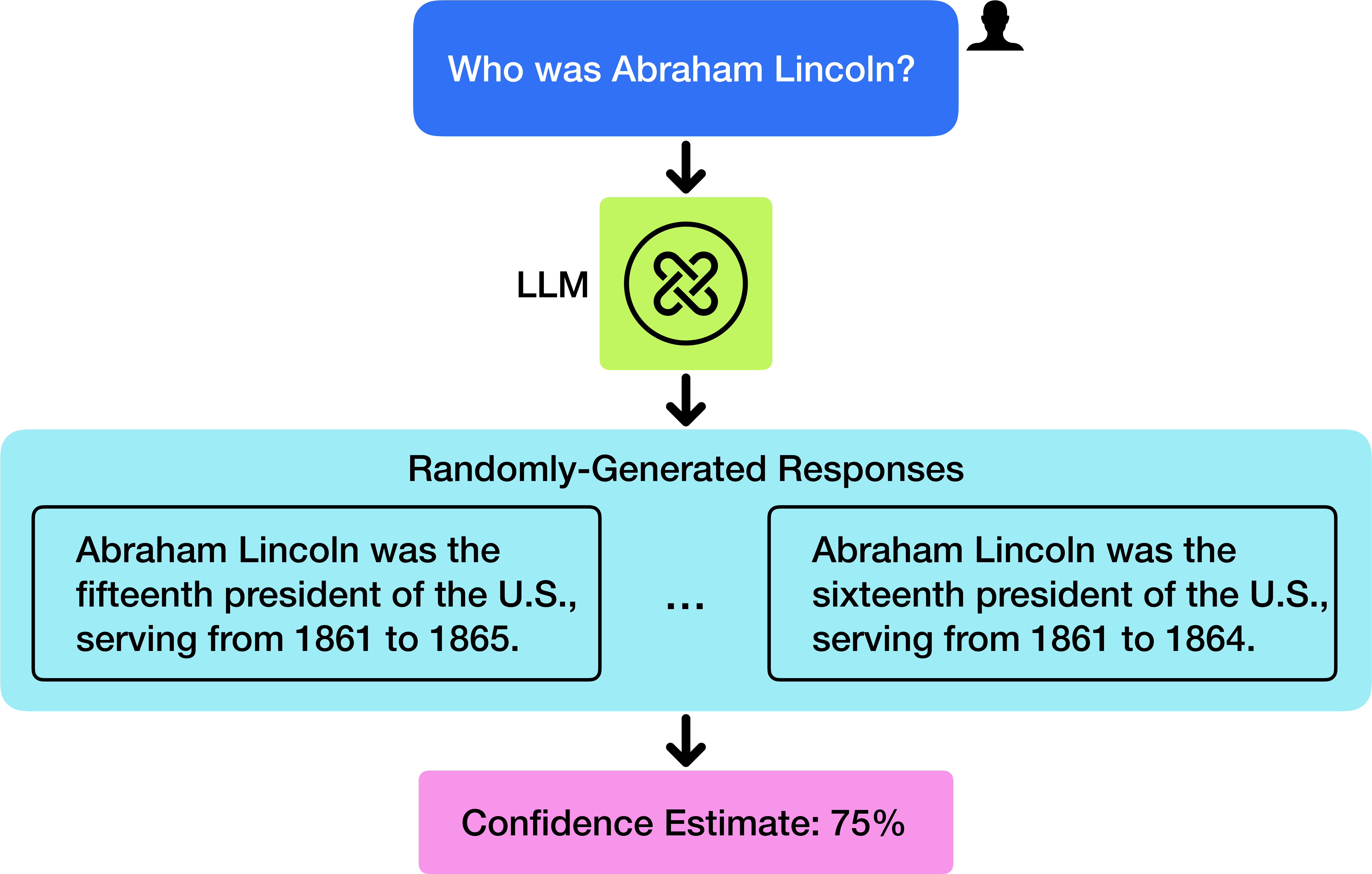}
        \caption{Black-box uncertainty quantification methods do not access the internal states or probabilities computed by the model, quantifying the model's uncertainty entirely from its natural-language response.}
        \Description[Black-box uncertainty quantification methods do not access the internal states or probabilities computed by the model, quantifying the model's uncertainty entirely from its natural-language response.]{Black-box uncertainty quantification methods do not access the internal states or probabilities computed by the model, quantifying the model's uncertainty entirely from its natural-language response.}
        \label{fig:black_box_uq}
    \end{minipage}
\end{figure}

\begin{enumerate}
    \item \text{Average Token Log-Probability.} The average of the negative log-probability of the tokens, which captures the average confidence of the model \cite{manakul2023selfcheckgpt}, is given by: ${\mathrm{Average}(p) = -\frac{1}{L_{i}} \sum_{j} \log(p_{ij}),}$
    where sentence $i$ consists of $L_{i}$ tokens. Note that the value of this metric increases as the conditional probability distribution of each token decreases, signifying an decrease in the model's confidence. The average token probability is related to the product of the token probabilities. 
    
    \item \text{Perplexity.} The perplexity of a model's prediction represents the exponential of the average of the negative log-probability of the tokens which comprise the sentence (response) generated by the LLM \cite{fadeeva2024fact}. Perplexity is given by: ${\mathrm{Perplexity}(p) = \exp \left(-\frac{1}{L_{i}} \sum_{j} \log(p_{ij}) \right).}$
    
    \item \text{Maximum Token Log-Probability.} The maximum token log-probability captures the token with the lowest conditional probability, which is given by: ${\mathrm{Maximum}(p) = \max_{j} -\log(p_{ij}).}$
    
    \item \text{Response Improbability.}
    This metric entails computing the probability of a given sentence given the conditional distribution for each token \cite{fadeeva2024fact}, where the probability distribution is conditioned on preceding tokens, and subtracting the resulting value from one. The uncertainty metric is defined as: ${\mathrm{Improb.} = 1 - \prod_{j} p_{ij}.}$
    
    \item \text{Entropy.} The maximum entropy of the probability distribution associated with each token can be utilized as a metric for UQ, given by: ${\mathrm{Entropy} = \max_{j} \mcal{H}(p_{j}),}$
    where $\mcal{H}$ represents the entropy of the probability distribution $p_{j}$ of token $j$. Some existing methods claim that this metric is better than the perplexity \cite{fadeeva2024fact}.
    Similarly, the predictive entropy \cite{malinin2020uncertainty} at input $x$ and output $y$ is defined as: ${\mcal{H}(Y \mid x) = - \int p(y \mid x) \ln p(y \mid x) dy.}$
    In the discrete case, the entropy associated with the output distribution of token $j$ in sentence $i$ is defined by: ${\mcal{H}_{ij} = - \sum_{w \in \mcal{D}} p_{ij}(w) \log p_{ij}(w),}$
    where $\mcal{D}$ denotes the dictionary containing all possible words in the model and $w$ represents a word in $\mcal{D}$.
    
\end{enumerate}

\paragraph{Black-Box UQ Methods}
In contrast, \emph{black-box} methods assume that the model's internal outputs cannot be accessed externally \cite{manakul2023selfcheckgpt, chen2023quantifying}. Hence, these methods quantify the uncertainty of the model entirely from the model's response to an input prompt. Prior work has discussed the pros and cons of both categories of UQ methods \cite{lin2023generating}. Concisely, white-box methods generally require access to the underlying architecture and intermediate outputs of an LLM, which is increasingly difficult to obtain given that many LLMs have become closed-source models, posing a significant limitation. In contrast, black-box models enable UQ of closed-source models such as OpenAI's GPT-4 \cite{achiam2023gpt} and Anthropic Claude \cite{anthropic2024claude}, which do not provide complete access to the model.
In general, black-box UQ methods for LLMs require the evaluation of the similarity between multiple responses generated by an LLM or an ensemble of LLMs on the same or similar prompts to quantify the uncertainty of the LLM on a given input prompt. Other black-box UQ methods, such as self-verbalized UQ methods, train the model to directly provide a natural-language estimate of its confidence. Here, we identify some prominent techniques for measuring the similarity between a pair of text fragments:

\begin{enumerate}
    \item \text{NLI Scores.}
    As described in Section~\ref{sec:nli_background}, NLI models, such as RoBERTa \cite{liu2019roberta} and DeBERTa \cite{he2020deberta}, classify the relationship between a pair of text fragment as either an entailment, a contradiction, or a neutral relation. 
    Many black-box methods utilize the probabilities (or logits) predicted by the NLI model for these three classes as a measure of the similarity between the two text fragments, which is ultimately used to quantify the uncertainty of the LLM. For example, given the probability $p_{\mathrm{entail}}$ predicted by an NLI model that a text fragment $t_{1}$ entails another text fragments $t_{2}$, we can define the similarity between both text fragments as: ${\mathrm{sim}(t_{1}, t_{2}) = p_{\mathrm{entail}}}$. Conversely, given the probability of contradiction $p_{\mathrm{contradict}}$, we can define the similarity between $t_{1}$ and $t_{2}$ as: ${\mathrm{sim}(t_{1}, t_{2}) = 1 - p_{\mathrm{contradict}}}$.

    \item \text{Jaccard Index.}
    The Jaccard index, also referred to as Intersection-over-Union measures the similarity between two sets by computing the ratio of the intersection of both sets and the union of both sets. Hence, the Jaccard index $J$ between two sets $\mcal{T}_{1}$ and $\mcal{T}_{2}$, where each set consists of the words that make up its associated text fragment, is given by: ${J(\mcal{T}_{1}, \mcal{T}_{2}) = \frac{\vert \mcal{T}_{1} \cap \mcal{T}_{2} \vert}{\vert \mcal{T}_{1} \cup \mcal{T}_{2} \vert}.}$
    Although the Jaccard index always lies between $0$ and $1$, making it a suitable metric \cite{pilehvar2013align, cronin2017comparison, qurashi2020document}, the Jaccard index does not consider the context of the text fragments, which is important in evaluating the similarity between both text fragments.

    \item \text{Sentence-Embedding-Based Similarity.}
    The similarity between two text fragments can also be determined by computing the cosine-similarity between the sentence embeddings associated with each text fragment. Sentence-embedding models transform natural-language inputs (or tokens) into a vector space, enabling direct computation of the similarity between two sentences (phrases). For example, Sentence-BERT (SBERT) \cite{reimers2019sentence} builds upon the pretrained BERT architecture to train a model that computes semantically-relevant sentence embeddings. Other similar models include LaBSE \cite{feng2020language} and SONAR \cite{duquenne2023sonar}. Since the sentence embeddings capture the context of the text fragment, this approach is less susceptible to the challenges faced by the Jaccard index, such as those that arise with negated words.
    
    \item \text{BERTScore.}
    The BERTScore \cite{zhang2019bertscore} measures the similarity between two sentences by computing the cosine-similarity between the contextual embedding of each token (word) in the reference sentence $t_{r}$ and the contextual embedding of the associated token in the candidate sentence $t_{c}$. The token embeddings are generated from NLI models to capture the context of the sentence. As a result, a given word might have different embeddings, depending on the context of the sentence in which it is used, addressing the challenges faced by the Jaccard similarity metric and word-embedding-based metrics. The BERTScore is composed of a precision $P_{\mathrm{BS}}$, recall $R_{\mathrm{BS}}$, and F1 $F_{\mathrm{BS}}$ score, given by:
    \begin{equation}
        P_{\mathrm{BS}} = \frac{1}{\vert t_{c} \vert} \sum_{\hat{w}_{j} \in t_{c}} \max_{w_{i} \in t_{r}} w_{i}^{\top} \hat{w}_{j}, \enspace
        R_{\mathrm{BS}} = \frac{1}{\vert t_{r} \vert} \sum_{w_{i} \in t_{r}} \max_{\hat{w}_{j} \in t_{c}} w_{i}^{\top} \hat{w}_{j}, \enspace
        F_{\mathrm{BS}} = 2 \frac{P_{\mathrm{BS}} \cdot R_{\mathrm{BS}}}{P_{\mathrm{BS}} + R_{\mathrm{BS}}},
    \end{equation}
    where each token in the candidate sentence is matched to its most similar token in the reference sentence. The BERTScore is obtained by computing the cosine-similarity between matched pairs. Since each token embedding is normalized, the cosine-similarity between a pair of embeddings simplifies to the inner-product. The recall score is related to the ROUGE metric \cite{lin2004rouge} used in evaluating text summaries and more broadly to the BARTScore \cite{yuan2021bartscore}. However, the ROUGE metric utilizes human-provided summaries as the reference.
\end{enumerate}

In the following sections, we describe the main categories of UQ methods for LLMs in detail, namely: (1) Token-Level UQ Methods; (2) Self-Verbalized UQ Methods; (3) Semantic-Similarity UQ Methods; and (4) Mechanistic Interpretability, outlined in \Cref{fig:outline}. Although mechanistic interpretability has not been widely applied to uncertainty quantification, we believe that insights from mechanistic interpretability can be more extensively applied to the uncertainty quantification of LLMs; hence, we include these methods within our taxonomy.

\section{Token-Level UQ}
\label{sec:token_level_uq}

We recall that the outputs of an LLM are generated by sampling from a probability distribution over the tokens that make up the outputs, conditioned on the preceding tokens in the outputs (see \Cref{{sec:background_uq_llm}}). Token-level UQ methods leverage the probability distribution over each token to estimate the probability of generating a given response from an LLM. Although a high predicted probability in a particular generation may not be indicative of its correctness over another, direct quantification of the model's generating distribution may lead to better understanding of the stochasticity of generations. 
Token-level UQ methods utilize the white-box UQ metrics discussed in \Cref{par:white_box_uq_llm} to estimate the randomness in the probability distribution associated with an LLM's response.
For example, some token-level UQ methods compute the entropy of the underlying probability distribution over the tokens \cite{xiao2021hallucination, ling2024uncertainty} or semantic clusters \cite{kuhn2023semantic} (referred to as semantic entropy) to estimate the LLM's confidence.
Likewise, a variant of SelfCheckGPT \cite{manakul2023selfcheckgpt} trains an $n$-gram model using multiple samples of the response of an LLM to a given query including its main response. Subsequently, SelfCheckGPT estimates the LLM's uncertainty by computing the average of the log-probabilities of the tokens generated by the $n$-gram model, given the original response of the LLM. Moreover, SelfCheckGPT proposes using the maximum of the negative log-probability to estimate the LLM's uncertainty. 

Token-based UQ methods generally perform poorly with long-form responses, because the product of the token probabilities decrease with longer responses, even when the responses are semantically equivalent to a shorter response. To address this limitation, token-based UQ methods employ a length-normalized scoring function  \cite{thomas2006elements, malinin2020uncertainty}, to reduce the dependence of the UQ metrics on the length of the sequence, given by: ${\mathrm{Product(p)} = \prod_{j = 1}^{L_{i}} p_{ij}^{\frac{1}{L_{i}}}}$,
where $L_{i}$ denotes the length of sentence $i$, and $p_{ij}$ is the conditional probability of token $j$, given all preceding tokens, in sentence $i$. The work in \cite{bakman2024mars} introduces Meaning-Aware Response Scoring (MARS) as an alternative to length-normalized scoring. MARS utilizes an importance function to assign weights to each token based on its contribution to the meaning of the sentence. The contribution of each token to the meaning of the sentence is determined using BEM \cite{bulian2022tomayto}, a question-answer evaluation model.
Taking a different approach, Claim-Conditioned Probability (CCP) \cite{fadeeva2024fact} decomposes the output of an LLM into a set of claims and computes the token-level uncertainty of each claim from its constituent tokens. CCP utilizes the OpenAI Chat API \cite{brown2020language, achiam2023gpt} to identify the main claims in a given response. By examining the component claims, CCP provides finer-grained uncertainty quantification compared to other UQ methods for LLMs.

As described, token-level UQ methods estimate the uncertainty of an LLM based on the conditional distribution associated with each token. Although this approach is effective in general, the conditional distribution of the tokens can be misleading in certain scenarios, especially when an initial token is incorrect but all the succeeding tokens are highly probable given the initial token. 
Trainable attention-based dependency (TAD) \cite{vazhentsev2024unconditional} trains a regression model on the conditional dependence between the tokens and applies the predicted factors to improve the estimated uncertainty of the LLM.
Lastly, we present token-level UQ methods that use specific prompting strategies to estimate the model's confidence. The work in \cite{kadavath2022language} shows that token-based UQ methods can be particularly effective in estimating the confidence of LLMs when the model is prompted to select an option when given a multiple-choice question. Specifically, the authors show that the model's probability distribution over the options in the prompt is well-calibrated, when presented with multiple-choice problems or problems with a True/False answer. Further, the authors fine-tune an LLM with a value head to predict the probability that the model knows the answer to a given question for each token. The probability associated with the LLM's final token is defined as the confidence of the LLM in its response for the given prompt. The results demonstrate that the LLM predictions of these probability values are well-calibrated, with an improvement in the calibration performance with larger models. Other follow-on work leveraging multiple-choice problems to estimate the uncertainty of LLMs includes \cite{ren2023a}.

\section{Self-Verbalized UQ}
\label{sec:self_verbalized_uq}
\begin{figure}[th]
    \centering
    \begin{minipage}[b][][b]{.48\textwidth}
        \centering
        \includegraphics[width=\columnwidth]{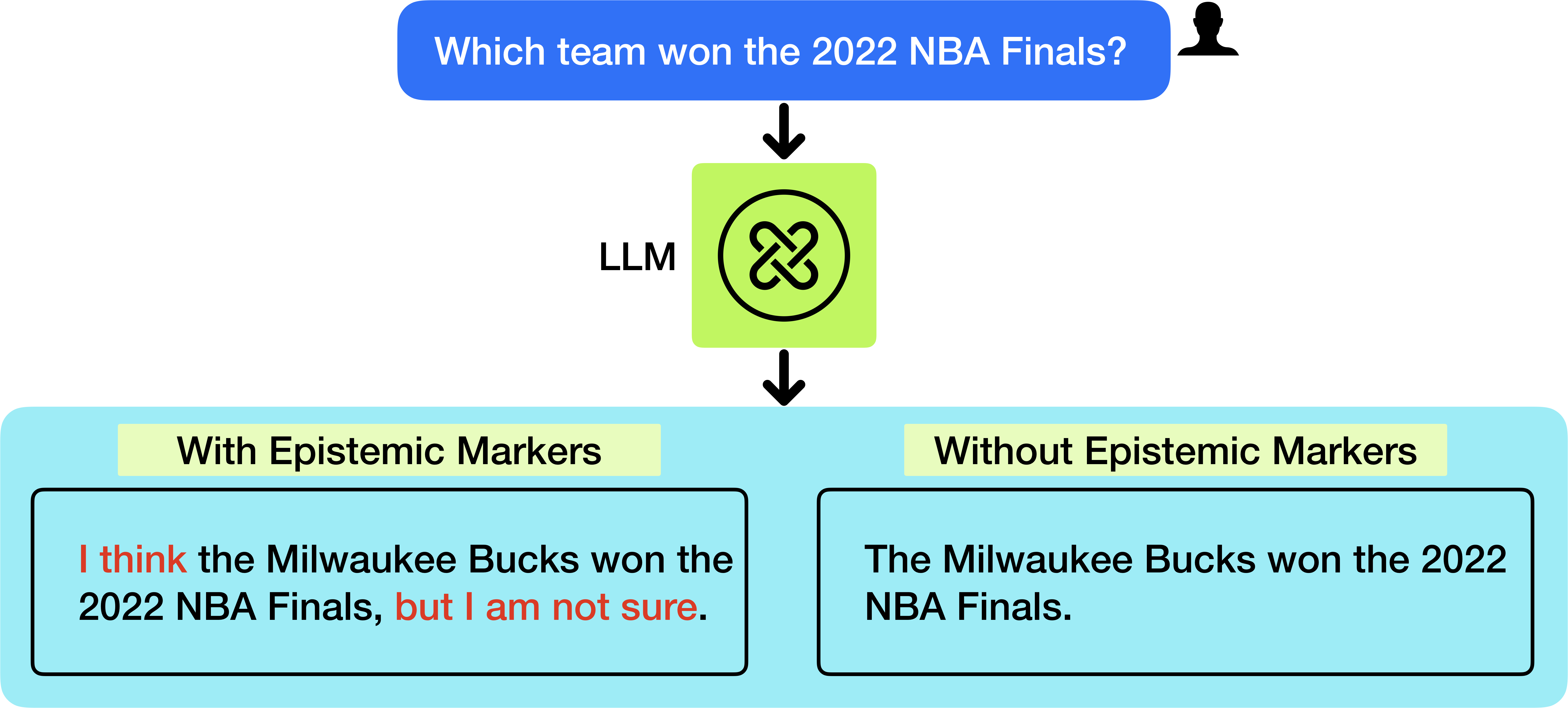}
        \caption{The LLM provides an incorrect response, but communicates its uncertainty using epistemic markers, e.g., ``I think."}
        \Description[The LLM provides an incorrect response, but communicates its uncertainty using epistemic markers, e.g., ``I think."]{The LLM provides an incorrect response, but communicates its uncertainty using epistemic markers, e.g., ``I think."}
        \label{fig:epistemic_uncertainty_markers}
    \end{minipage}%
    \hfill %
    \begin{minipage}[b][][b]{.48\textwidth}
        \centering
        \includegraphics[width=\columnwidth]{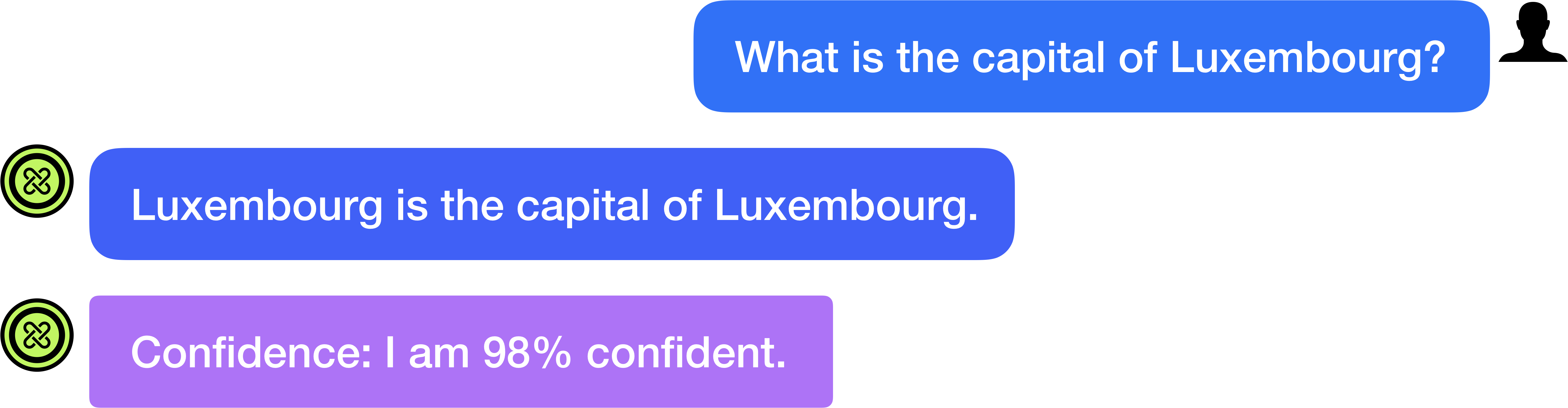}
        \caption{LLMs can be trained or fine-tuned to provide numeric estimates of their confidence in the factuality of their response.}
        \Description[LLMs can be trained or fine-tuned to provide numeric estimates of their confidence in the factuality of their response.]{LLMs can be trained or fine-tuned to provide numeric estimates of their confidence in the factuality of their response.}
        \label{fig:self_verbalized_uq_prob}
    \end{minipage}
\end{figure}

Self-verbalized uncertainty quantification methods seek to harness the impressive learning and reasoning capabilities of LLMs to enable an LLM to express its confidence in a given response through natural-language. Self-verbalized uncertainty estimates (e.g., expressed as probabilities) are more easily interpretable to humans, especially when the estimates are provided using widely-used epistemic uncertainty markers \cite{tang2024evaluation, yona2024can}, e.g., words like \emph{I am not sure...} or \emph{This response might be...} \Cref{fig:epistemic_uncertainty_markers} illustrates the use of epistemic markers by an LLM to convey its uncertainty, when asked of the team that won the 2022 NBA Finals. The response of the LLM is actually incorrect; however, by expressing its uncertainty, a user may be more inclined to verify the factuality of the LLM's response. Prior work has shown that LLMs typically fail to accurately express their confidence in a given response, often using decisive words that suggest confidence, while at the same time being unsure of the accuracy of their response. Empirical studies \cite{krause2023confidently} have shown that poor calibration of LLM's self-verbalized confidence estimates is more pronounced in low-data language settings, e.g., Hindi and Amharic.

To address this challenge, prior work in \cite{mielke2022reducing} trains a learned model (calibrator) that predicts the probability that an LLM's response to a given prompt is correct, given the input prompt, its response, and the LLM's representations of the prompt and its response. In addition, the output of the calibrator and the LLM's original response are subsequently used in fine-tuning a generative model \cite{smith2020controlling} to produce a linguistically calibrated response, aligning the verbal expression of the LLM's confidence with its probability of factual correctness. 
However, the resulting verbalized uncertainty lacks a numerical value, making it difficult for users to assess the relative confidence of the LLM. Follow-on work in \cite{lin2022teaching} introduces the notion of \emph{verbalized probability}, providing a definite numerical value of the model's confidence, e.g., in \Cref{fig:self_verbalized_uq_prob}, or a scaled characterization of the model's confidence in words, e.g., \emph{low}, \emph{medium}, or \emph{high} confidence. The authors of \cite{lin2022teaching} fine-tune GPT-3 on their proposed CalibratedMath benchmark dataset using supervised learning, demonstrating that the verbalized probability generalizes well; however, best performance is achieved in in-distribution scenarios.

\begin{figure}[th]
    \centering
    \includegraphics[width=0.5\textwidth]{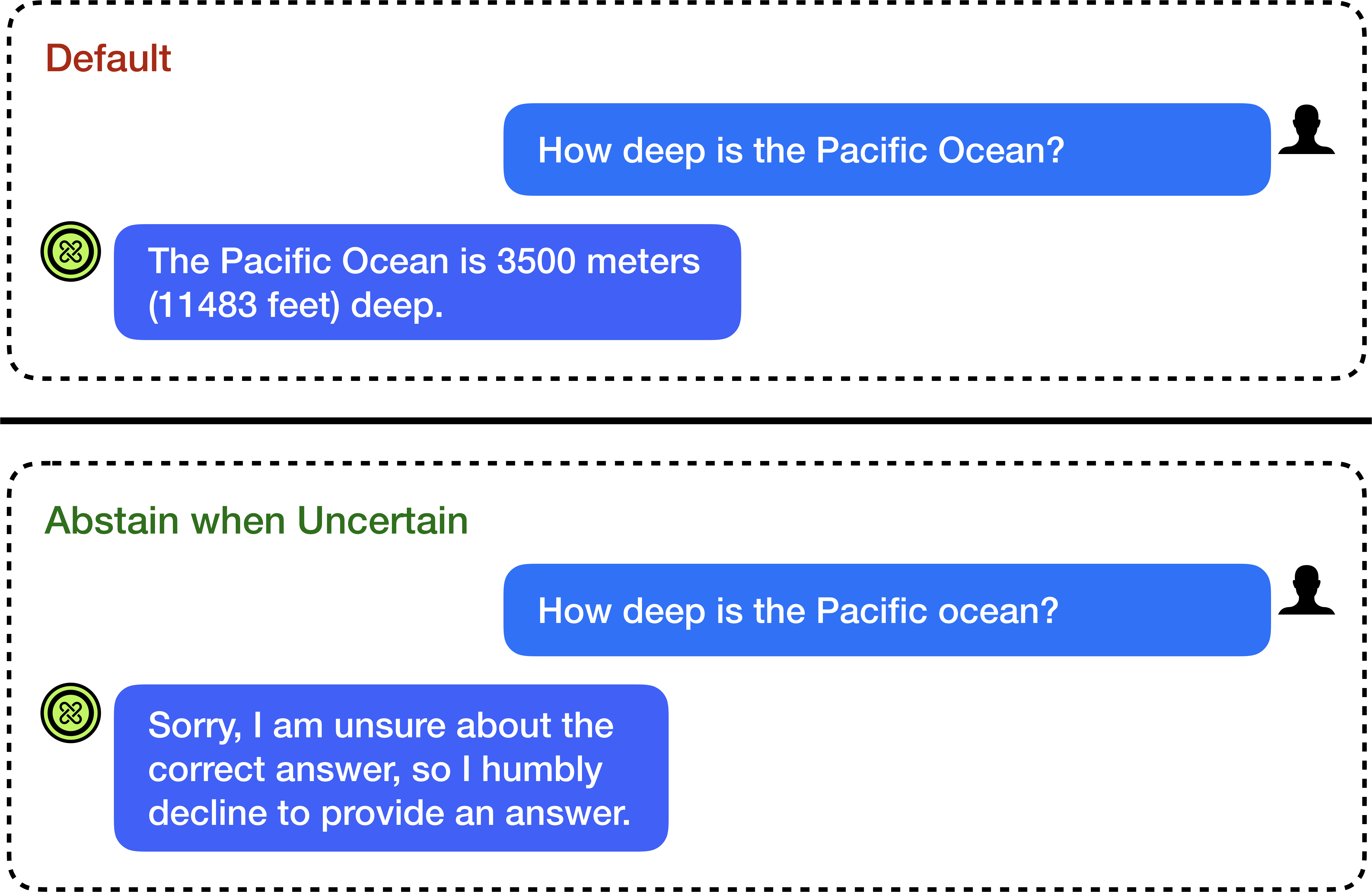}
    \caption{Some self-verbalized UQ methods fine-tune an LLM to refrain from answering when it is uncertain about the answer.}
    \Description[Some self-verbalized UQ methods fine-tune an LLM to refrain from answering when it is uncertain about the answer.]{Some self-verbalized UQ methods fine-tune an LLM to refrain from answering when it is uncertain about the answer.}
    \label{fig:self_verbalized_uq_honesty}
\end{figure}

More recent work has investigated other training approaches for fine-tuning LLMs to accurately express their confidence verbally.
LACIE \cite{stengel2024lacie} introduces a two-agent speaker-listener architecture to generate training data for fine-tuning an LLM, where the reward signal is a function of the ground-truth answer and the listener's perceived confidence of the speaker's response. In essence, LACIE aims to fine-tune an LLM to produce a response composed of epistemic markers that are aligned with the model's confidence in the correctness of its response.  
Likewise, the work in \cite{yang2024can} proposes a knowledge-transfer training architecture where the knowledge from a bigger LLM (the teacher), e.g., GPT-4 \cite{achiam2023gpt}, is distilled into a smaller LLM (the student), e.g., Vicuna-7B \cite{chiang2023vicuna}, using chain-of-thought reasoning. The student LLM is fine-tuned to provide its confidence (expressed as a value between $0$ and $100$) along with its response to an input prompt.
A line of existing work \cite{xu2024sayself, tao2024trust} utilizes reinforcement learning to fine-tune an LLM to improve the alignment of the confidence estimates expressed by the LLM with its factual accuracy. While SaySelf \cite{xu2024sayself} relies on self-reflective rationales to improve the calibration of the verbalized confidence, the work in \cite{tao2024trust} uses reinforcement learning from human feedback (RLHF) to define a reward function consisting of a quality component in addition to an alignment component. Similarly, the work in \cite{band2024linguistic} fine-tunes Llama~2 \cite{touvron2023llama} using supervised learning and reinforcement learning, to produce calibrated verbalized confidence estimates that enable a user to make informed decisions on related questions.
Lastly, other recent work, e.g., \cite{yang2023alignment, feng2024don}, seeks to fine-tune LLMs to abstain from providing an answer to a question when faced with doubt \cite{tomani2024uncertainty}, which is illustrated in \Cref{fig:self_verbalized_uq_honesty}.

Despite these efforts, in many cases, LLMs still fail to accurately express their confidence verbally \cite{xiong2023can, groot2024overconfidence}, typically exhibiting overconfidence, with confidence values primarily between 80\% and 100\%, often in multiples of $5$, similar to the way humans interact. This weakness decreases with the size of an LLM. Nonetheless, large-scale LLMs still display overconfidence, albeit at a smaller rate. However, effective prompting strategies to reduce the calibration error of these models exist. Although verbalized confidence estimates are better calibrated than raw, conditional token probabilities \cite{tian2023just}, existing empirical studies \cite{ni2024large} suggest that token-based UQ methods generally yield better-calibrated uncertainty estimates compared to their self-verbalized UQ counterparts.

\section{Semantic-Similarity UQ}
\label{sec:semantic_similarity_uq}

\begin{figure}
    \centering
    \includegraphics[width=0.45\linewidth]{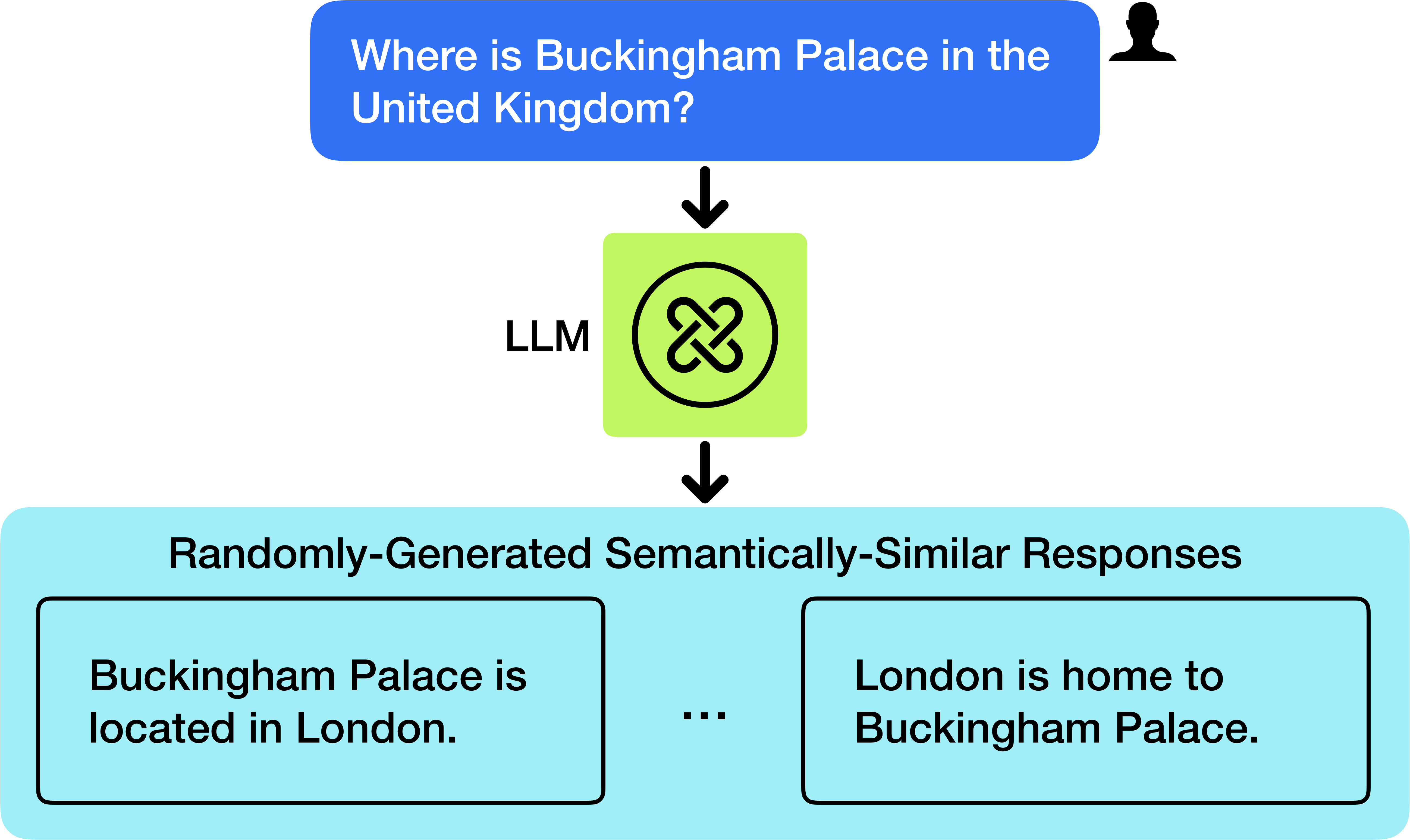}
    \caption{When prompted to answer a question, e.g., ``Where is Buckingham Palace in the United Kingdom?'', an LLM might generate many variations of the same sentence. Although the form of each response may differ at the token-level, the semantic meaning of the sentences remains consistent. Semantic-similarity UQ techniques exploit semantic clustering to derive UQ methods that are robust to these variations in the form of the responses.}
    \Description[When prompted to answer a question, e.g., ``Where is Buckingham Palace in the United Kingdom?'', an LLM might generate many variations of the same sentence. Although the form of each response may differ at the token-level, the semantic meaning of the sentences remains consistent. Semantic-similarity UQ techniques exploit semantic clustering to derive UQ methods that are robust to these variations in the form of the responses.]{When prompted to answer a question, e.g., ``Where is Buckingham Palace in the United Kingdom?'', an LLM might generate many variations of the same sentence. Although the form of each response may differ at the token-level, the semantic meaning of the sentences remains consistent. Semantic-similarity UQ techniques exploit semantic clustering to derive UQ methods that are robust to these variations in the form of the responses.}
    \label{fig:semantically_similar_responses}
\end{figure}

Semantic-similarity uncertainty quantification methods examine the similarity between multiple responses of an LLM to the same query \cite{kuhn2023semantic, chen2023quantifying, lin2023generating} by focusing on the \textit{meaning} (i.e., the semantic content of a generated sentence) rather than the \textit{form} (i.e., the string of tokens that the model predicts) \cite{kuhn2023semantic} of the responses. For example, consider the prompt (question) given to an LLM: \emph{Where is Buckingham Palace in the United Kingdom?} Standard sampling from an LLM can produce many variations of the same answer when prompted with this question, as illustrated in \Cref{fig:semantically_similar_responses}. However, while an LLM may be uncertain about which sequence the user may prefer, most variations do not alter the meaning of the sentence. This difference in the ordering of the tokens in each response may lead to different token probabilities, which in turn may negatively impact the accuracy of other uncertainty quantification methods, such as token-level UQ methods.

Since semantic similarity is a relative metric, its outputs are in general model-dependent, posing a central challenge. A recent line of work uses NLI models, such as RoBERTa \cite{liu2019roberta} and DeBERTa \cite{he2020deberta} (discussed in \Cref{sec:nli_background}),  to compute entailment probabilities. The work in \cite{aichberger2024semantically} proposes upweighting tokens that have large gradients with respect to the NLI model to maximize the probability of contradiction to generate semantically-varied responses. In addition, the method in \cite{tanneru2024quantifying} proposes using a chain-of-thought agreement (CoTA) metric that uses entailment probabilities to evaluate the agreement between CoT generations, concluding that CoTA semantic uncertainty leads to more robust model faithfulness estimates than either self-verbalized or token-level uncertainty estimates. The authors of \cite{chen2023quantifying} propose using a combined measure of confidence that incorporates entailment probabilities along with a verbalized confidence score, and selects the generation with the highest confidence. The UQ method in \cite{becker2024cycles} proposes generating multiple explanations for each plausible response and then summing the entailment probabilities. Another work \cite{kossen2024semantic} introduces semantic entropy probes, wherein semantic clusters are grown iteratively using entailment probabilities. Each new generation is either added to an existing cluster if entailment holds, or added to a new cluster. Then, a linear classifier is trained to predict high-entropy prompts. Furthermore, the method in \cite{martin2022facter} uses a database of user-verified false statements to build a semi-automated fact-checking system that uses entailment probabilities with database queries as a metric for confidence in a statement's falseness. 

In addition to using NLI models to evaluate factual similarities between responses, some methods use language embeddings \cite{petukhova2024text} to cluster responses based on their semantic similarity and reason about uncertainty over the clusters, e.g., semantic density in \cite{qiu2024semantic}. First, several reference responses are generated by sampling the model. Then, the overall uncertainty per response is computed using the entailment scores, taking values in the set $\{0, 0.5, 1\}$. The semantic density is then used to accept or reject a target response based on the similarity to the target responses.  The supervised approach in \cite{he2024mitigating} utilizes the K-means algorithm to first cluster synonyms, which are attended by the LLM during training. The work in \cite{hu2024enhancing} introduces a method to achieve semantically-aligned item identification embeddings based on item descriptions, which aid in aligning LLM-based recommender systems with semantically-similar generations when item descriptions are sparse. Further, the method in \cite{wang2024clue} prompts an LLM to generate concepts (effectively semantic clusters) and uses an NLI-based concept scorer along with the entropy over the concepts to quantify the overall uncertainty of the LLM. ClusterLLM \cite{zhang2023clusterllm} uses a frozen instruction-trained LLM to guide clustering based on triplet queries (e.g., does A match B better than C?), achieving more semantically-aligned embeddings.

However, assigning responses to a single cluster precludes assignment to another, when in reality a response may belong to more than one class, limiting the effectiveness of clustering-based semantic-similarity UQ methods. To address this challenge, another line of work extends clustering-based methods to graphs, which may express the complex relationship between responses more explicitly. The work in \cite{ao2024css} proposes Contrastive Semantic Similarity, which uses responses as vertices and CLIP cosine similarities as edges. The overall uncertainty is computed from the eigenvalues of the graph Laplacian, and the eigenvectors can be used to assign clusters more effectively. Similarly, the approach in \cite{da2024llm} uses edges weights determined directly from NLI models and extends the graph-Laplacian-based uncertainty metric to include additional semantic uncertainty, such as Jaccard similarity.
The authors of \cite{jiang2024graph} introduce a claim-and-response structure wherein edges are added between a claim and response if the response entails the claim. The centrality metrics are used to estimate per-claim uncertainty and integrate low-uncertainty claims into further generations. In addition, Kernel Language Entropy \cite{nikitin2024kernel} clusters responses to construct a kernelized graph Laplacian, which is used to estimate fine-grained differences between responses in a cluster. 

A few works that learn to estimate semantic meanings without NLI models using supervised approaches have also been proposed. In \cite{liu2024uncertainty}, the authors use an auxiliary tool LLM to compute a similarity score between the target LLM's generation and the tool LLM's generation and learns an uncertainty estimation function to estimate the similarity score. In \cite{jung2024trust}, the authors propose a cascading chain of increasingly complex LLM judges to evaluate the predecessor's preference between two generations. A calibration dataset is used to learn a threshold that determines each judge's minimum confidence level. The confidence thresholds are tuned in order to guarantee that the appropriate judge is selected to generate a satisfactory response.

\section{Mechanistic Interpretability}
\label{sec:mechanistic_interpretability_uq}

Mechanistic interpretability (MI) aims to understand the inner workings of LLMs to pinpoint the potential sources of uncertainty, by uncovering causal relationships~\cite{bereska2024mechanistic}. Several survey papers have provided a taxonomy of mechanistic interpretability in the field of transformer-based language models~\cite{rai2024practical}, focused on AI safety~\cite{bereska2024mechanistic} or interpretability of language models in general~\cite{zhao2024explainability}.
\begin{figure}[th]
    \centering
    \includegraphics[width=0.55\linewidth]{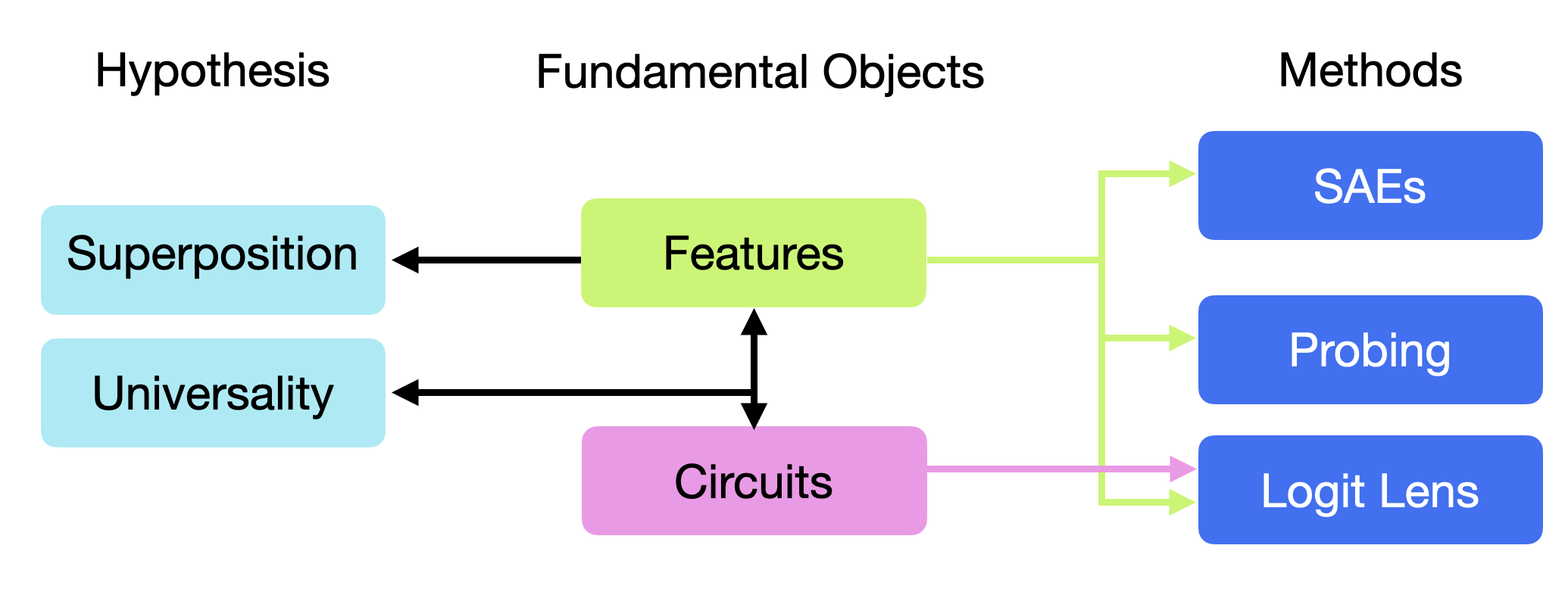}
    \caption{Taxonomy of Mechanistic Interpretability~\cite{rai2024practical}.}
    \Description[Taxonomy of Mechanistic Interpretability~\cite{rai2024practical}.]{Taxonomy of Mechanistic Interpretability~\cite{rai2024practical}.}
    \label{fig:MI-taxonomy}
\end{figure}

We start by discussing a few key concepts of mechanistic interpretability, summarized in Figure~\ref{fig:MI-taxonomy}.  Features are the unit for encoding knowledge in a neural network. For example, a neuron or set of neurons consistently activating for Golden Gate Bridge can be interpreted as the ``Golden Gate Bridge'' feature~\cite{Templeton2024Scaling}.
\emph{Superposition} \cite{elhage2022toy} is often a key hypothesis in mechanistic interpretability~\cite{bereska2024mechanistic}, due to the fact that the same neuron seems to activate in multiple, distinct contexts, a phenomenon known as polysemanticity~\cite{cunningham2023sparse}. The superposition hypothesis claims that the set of $N$ neurons encode $M>N$ features, by allocating each feature to a linear combination of neurons, which are in almost orthogonal directions, leading to an overcomplete set of basis. On the other hand, the work in~\cite{engels2024not} suggests that there exists circular features corresponding to days of the week and months of the year, breaking the assumption that high-level features are linearly represented in the activation space.
Circuits, another fundamental concept, refers to sub-graphs of the network that consist of features and weights connecting them. Recent research have aimed to perform comprehensive circuit analysis on LLMs in order to construct a full mapping from specific circuits to functionalities of the language model~\cite{lieberum2024gemma, dunefsky2024transcoders}.
The hypothesis of universality, related to both features and circuits, claims that similar features and circuits exist across different LLMs.

Methods in MI can be broadly classified into the following categories: logit lens, probing, and sparse auto-encoders methods, each discussed briefly. 
Logit lens methods project the activations from various layers of the LLM back into the vocabulary space, allowing for interpreting intermediate predictions and information encoded in activations~\cite{geva2020transformer, lieberum2023does}.
Probing methods aim to find which intermediate activations encode specific information (e.g., syntactic, semantic, or factual knowledge), by training a linear classifier as a \textit{probe} to predict the existence of a certain feature~\cite{belinkov2022probing, gurnee2023finding}. Despite being simple and successful, probing methods only reveal correlations instead of causal relations, limiting their use in MI.
Sparse auto-encoders (SAEs) represent a popular architecture applied in MI to directly identify meaningful feature activations within LLMs and the causal relations between them. SAEs map the feature vectors onto a much higher dimensional space with strong sparsity, in order to disentangle the features that were in superposition. In these methods, an encoder-decoder pair $(z,\hat x)$ is trained to map $\hat x(z(x))$ back to the model's activation $x$, given by: 
${z = \sigma( W_\text{enc}x+ b_\text{enc})}$, ${{\hat x} =  W_\text{dec} z+ b_\text{dec}}$.
The specific implementation of the activation function can vary, with a common choice of the activation function given by the ReLU \cite{dunefsky2024transcoders, cunningham2023sparse}. In \cite{gao2024scaling}, ${\sigma = \mathrm{TopK}}$ is used to keep only the $k$-largest latents, simplifying tuning and outperforming ReLU. In ~\cite{lieberum2024gemma}, ${\sigma = \mathrm{JumpReLU}}$ is chosen due to its slightly better performance and the ability to allow for a variable number of active latents at different tokens. 
In~\cite{dunefsky2024transcoders}, the authors train the architecture differently with transcoders, where the faithfulness term in the loss function measures the error between the output and the original MLP sub-layer output, instead of the original input.
In \cite{yun2021transformer}, the authors hypothesize that contextualized word embeddings are linear superpositions of transformer factors. For example,  the word ``apple" can be decomposed into: ${\mathrm{apple} =0.09 \mathrm{dessert} + 0.11 \mathrm{organism} + 0.16 \mathrm{fruit} + 0.22 \mathrm{mobile\&IT} + 0.42 \mathrm{other}}$.
The authors aim to learn a comprehensive dictionary of word factors. In doing so, they distinguish between low, mid, and high-level factors by looking at the change in the importance score across layers. Low-level factors correspond to word-level polysemy disambiguation; mid-level factors are sentence-level pattern formation; and high-level factors correspond to long-range dependency, which have to be manually distinguishable from mid-level factors, although it could be done with black-box interpretation algorithms as well.
In~\cite{tamkin2023codebook}, the authors quantize features into sparse ``codebook'' features, providing the capability to control the network behavior.

Prior work has employed techniques from mechanistic interpretability  to track the progress of models during training \cite{nanda2023progress}, to explain the outputs of models \cite{schwab2019cxplain}, and to improve the accuracy of LLMs \cite{burns2022discovering}. The work in \cite{burns2022discovering} demonstrates that the accuracy of the latent knowledge of LLMs is less sensitive to the input prompts, with its accuracy remaining relatively constant even when the LLM is prompted to generate incorrect responses. Likewise, ReDeEP \cite{sun2024redeep} examines the latent knowledge of an LLM to decouple the effects of external knowledge from knowledge bases and the internal knowledge in the model on hallucinations in retrieval-augmented generation. 
Further, prior work has examined hallucinations in LLMs through the lens of mechanistic interpretability \cite{yu2024mechanisms, wang2024latent}. The work in \cite{yu2024mechanisms} investigates the role of an LLM's hidden states in contributing to hallucinations, quantifying the contributions of lower-layer and upper-layer MLPs and attention heads to factual errors. In addition, the method in \cite{ferrando2024know} leverages mechanistic interpretability to identify the boundaries of an LLM's internal knowledge of its own capabilities, which could be used to prevent a model from answering questions on certain subjects (i.e., in safeguarding the model) or to prevent hallucinations when the model does not know about certain subjects.  Lastly, the work in \cite{ahdritz2024distinguishing} trains small classifiers (linear and non-linear MLPs) on the activations of a small LLM to predict the uncertainty level of a larger LLM, demonstrating that the classifiers generalize to unseen distributions.
Although there is an inextricable link between understanding the inner workings of LLMs and quantifying their uncertainty when prompted by a user, the connections between mechanistic interpretability and uncertainty quantification have not been extensively explored. For example, certain neural activation patterns in LLMs might be associated with the expression of uncertainty by the model. In addition, when faced with doubt, an LLM might utilize certain features (words/concepts), that could be detected from its neural activations. Identifying the specific intermediate activations and features of an LLM that are relevant for uncertainty quantification remains an open research challenge.
We describe this open challenge in Section~\ref{ssec:challenges_mechanical interpretability}.

\section{Calibration of Uncertainty}
\label{sec:calibration}
\begin{figure}
    \centering
    \includegraphics[width=0.6\linewidth]{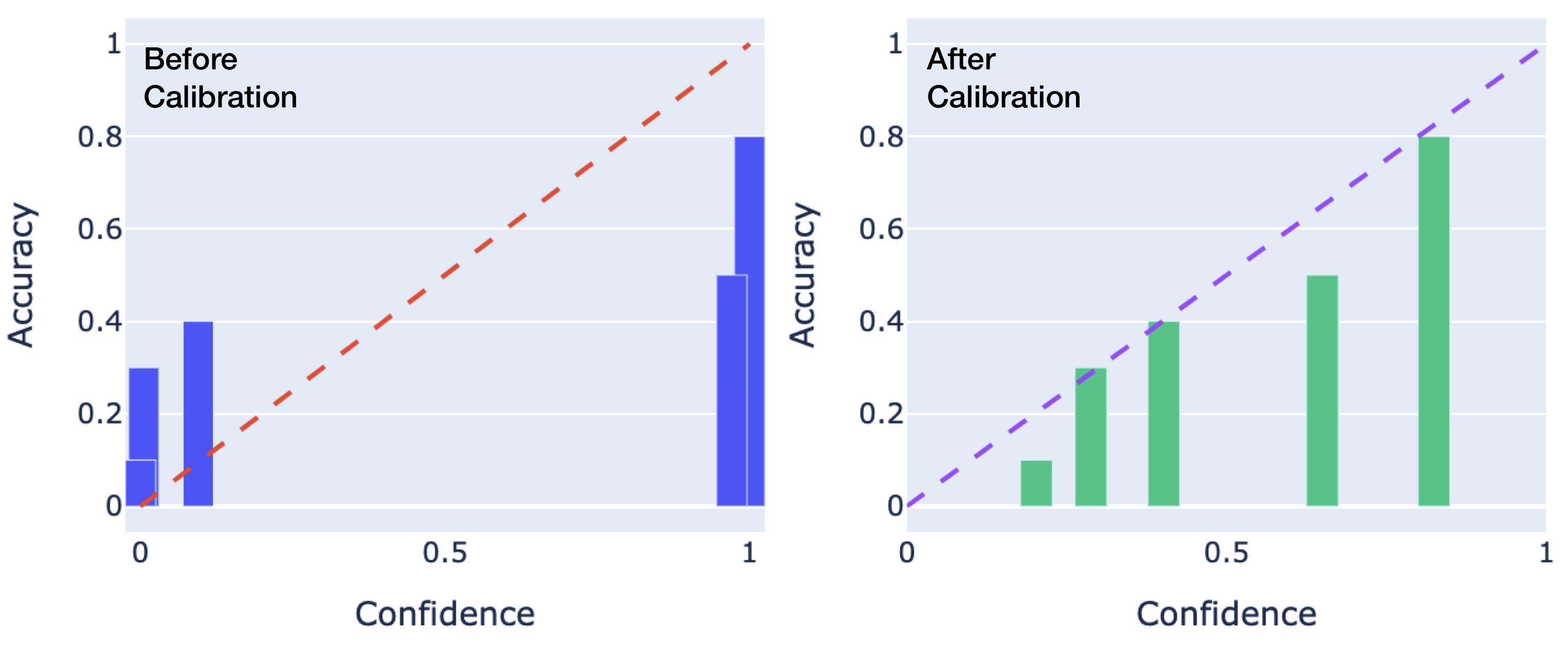}
    \caption{The confidence estimates provided by many UQ methods are not always calibrated, i.e., the observed frequencies do not match the estimates. Calibration techniques correct these confidence estimates for better alignment with the observed accuracy.}
    \Description[The confidence estimates provided by many UQ methods are not always calibrated, i.e., the observed frequencies do not match the estimates. Calibration techniques correct these confidence estimates for better alignment with the observed accuracy.]{The confidence estimates provided by many UQ methods are not always calibrated, i.e., the observed frequencies do not match the estimates. Calibration techniques correct these confidence estimates for better alignment with the observed accuracy.}
    \label{fig:calibration}
\end{figure}

In many cases, the confidence estimates computed by the UQ methods presented in the preceding sections are not well-calibrated i.e., aligned with the observed frequencies of the responses (accuracy of the model).
However, reliability of the confidence estimates of an LLM's output remains crucial to the safe deployment of LLMs. 
As a result, we would like the confidence estimates to be calibrated. Formally, for a perfectly-calibrated confidence estimate $p$, we have that, ${\forall p \in [0, 1]}$: 
\begin{equation}
    \label{eq:prob_calibration}
    \mbb{P}[Y = \hat{Y} \mid \hat{P} = p] = p,
\end{equation}
where $Y$ and $\hat{Y}$ represent random variables denoting the ground-truth and predicted outputs from the model, respectively, and $\hat{P}$ represents a random variable denoting the confidence associated with the predicted output $\hat{Y}$ \cite{guo2017calibration}. 
In \Cref{fig:calibration}, we show poorly-calibrated confidence estimates on the left, where the estimated confidence of the model is not well-aligned with the observed accuracy of the model. The dashed-line illustrates perfect alignment between the estimated confidence of the model and its accuracy. In this example, confidence estimates of the model above $0.5$ tend to be overconfident, exceeding the accuracy of the model. Conversely, confidence estimates that are less than $0.5$ tend to be underconfident. Calibration techniques improve the alignment of the estimated confidence of the model with the observed accuracy, with the estimated confidence more closely following the dashed-line, as shown on the right in \Cref{fig:calibration}.
We review some metrics for quantifying the calibration of a model's confidence estimates.

\subsubsection*{Expected Calibration Error (ECE)}
The Expected Calibration Error (ECE) measures the expected deviation between the left-hand side and right-hand side of \eqref{eq:prob_calibration}, with:
${\mbb{E}_{\hat{P}}\left[ \left\lvert \mbb{P}[Y = \hat{Y} \mid \hat{P} = p] - p \right\rvert \right],}$
where the expectation is taken over the random variable $\hat{P}$. Computing the expectation in the ECE is intractable in general. Hence, the work in \cite{naeini2015obtaining} introduces an approximation of the ECE, which partitions the confidence estimates into equal-width bins and computes the difference bin-wide, with: ${\mathrm{ECE} = \sum_{m = 1}^{M} \frac{\lvert B_{m} \rvert}{n} \left\lvert \acc(B_{m}) - \conf(B_{m}) \right\rvert,}$
where the confidence estimates are divided into $M$ bins with the i'th bin denoted by $B_{i}$, and $\acc$ and $\conf$ denote the average accuracy and confidence of the samples in a bin.

\subsubsection*{Maximum Calibration Error (MCE)}
Alternatively, we may seek to quantify the maximum deviation between the left-hand and right-hand sides of \eqref{eq:prob_calibration}, representing the worst-case error, which is often useful in safety-critical applications. The  Maximum Calibration Error (MCE) is given by:
${\max_{p \in [0, 1]} \left\lvert \mbb{P}[Y = \hat{Y} \mid \hat{P} = p] - p  \right\rvert,}$
which is also challenging to compute exactly, like the ECE. As a result, we can estimate an upper bound, given by:
${\mathrm{MCE} = \max_{m \in \{1,\ldots,M\}} \left\lvert \acc(B_{m}) - \conf(B_{m}) \right\rvert,}$
as introduced in \cite{naeini2015obtaining}.
Metrics for quantifying the calibration error of confidence estimates are further discussed in 
\cite{guo2017calibration, niculescu2005predicting, nixon2019measuring}.

We can categorize calibration techniques for uncertainty estimation as either \emph{training-based} or \emph{training-free} calibration methods.
Training-based calibration methods comprise supervised techniques that modify the network's weights and various types of self-verbalization, where the model qualifies and refines its outputs based on its own reasoning or feedback about uncertainty. In contrast, training-free calibration methods include statistical techniques that operate on a frozen learned model.

\subsection{Training-Free Calibration Methods}
Training-free calibration methods do not modify the weights of the model to produce calibrated predictions, e.g., Platt scaling \cite{platt1999probabilistic}, isotonic regression \cite{zadrozny2001obtaining, zadrozny2002transforming}, and conformal prediction \cite{shafer2008tutorial}.
Here, we discuss conformal prediction in greater detail. Conformal prediction (CP) is a powerful technique used to quantify the uncertainty of a model's predictions by providing prediction sets that are guaranteed to contain the true outcome with a specified probability. Given a prediction model $f$ and a calibration dataset $\mathcal D_{\text{cal}} = \{(x, y)_i)\}_{i=1}^N$, conformal prediction aims to compute a set of nonconformity scores $\mathcal S = \{(s)_i\}_{i=1}^N$ over $\mathcal D_{\text{cal}}$, which reflect how closely each prediction $f(x_i)$---such as the confidence estimate provided by the aforementioned UQ methods---aligns with the true label $y_i$. Given a coverage level $\hat \varepsilon$ (effectively a budget for incorrect predictions) and $\mathcal S$, CP aims to construct a prediction set $C(x_{n+1})$ for a new \textit{test} data point $x_{n+1}$:
${C(x_{n+1}) = \left\{ y : f(x_{n+1}) \leq q_{1-\hat \varepsilon}(s_1, s_2, \dots, s_n) \right\},}$
along with the probabilistic guarantee:
${\mathbb P (x_{n+1} \in C(x_{n+1}) | \mathcal D_{\text{cal}}) \geq 1-\varepsilon(\delta),}$
where $q_{1-\hat \varepsilon}$ is the $(1 - \hat \varepsilon)$-quantile of the nonconformity scores from the calibration set
and $\delta$ is a tunable failure probability associated with the randomness in sampling $\mathcal D_{\text{cal}}$. By applying a Hoeffding-style argument \cite{shafer2008tutorial}, one can show that $\varepsilon$ can be selected, e.g.,  using the cumulative distribution function of the Beta distribution: ${\varepsilon := \text{Beta}^{-1}_{N + 1 - v, v}(\delta), \quad v:=\lfloor (N+1)\hat \varepsilon \rfloor,}$
where $\hat \varepsilon$  is the target coverage level. 

Provided that the nonconformity scores represent the true conditional probabilities, conformal prediction produces the tightest prediction set that minimizes the number of false positives (i.e., maximizes the discriminative power) among all set-valued predictors such that the user-specified coverage level holds~\cite[Theorem~1]{sadinle2019least}. As a result,
LLMs that are calibrated with conformal prediction will have the smallest prediction sets on average, and therefore the least ambiguity in their responses. A number of papers employ conformal prediction for uncertainty quantification of LLMs, e.g., for semantic uncertainty quantification \cite{wang2024conu} and calibration \cite{liu2024multi}.
In addition to conformal prediction, information-theoretic approaches have been developed to manage and calibrate uncertainty in sequential decision-making processes \cite{zhao2022calibrating}, e.g., entropy-rate control and multicalibration \cite{detommaso2024multicalibration}, which involves grouping data points into subgroups and ensuring the model is calibrated with respect to each of these subgroups . A model can also be calibrated to control a heuristic estimate of risk, such as human agreement  \cite{jung2024trust} or Pareto-optimality of the response correctness \cite{zhao2024pareto}. 

\subsection{Training-Based Calibration Methods}
We can group training-based calibration techniques into ensemble-based calibration methods, few-shot calibration methods, and supervised calibration methods.

\subsubsection{Ensemble-Based Calibration}
Ensemble-based calibration (model ensembling) seeks to estimate uncertainty by querying many similar models (for example, the same architecture trained with different random seeds) and comparing their outputs. Prompt ensembles enhance calibration by combining the outputs of multiple prompts \cite{jiang2023calibrating}. One common and effective ensembling strategy involves utilizing the majority vote. Given $K$ models predicting a response $l_i$, the majority vote is selected as:
${P_{\text{acc}}(\hat{y} = l_i) = \sum_{k=1}^{K} P_k(\hat{y}_k = l_i) \mathbb{I}(\hat{y}_k = l_i).}$
The ensemble vote is then the response $l_i$ with the highest aggregate confidence. Another class of ensemble-based methods evaluates \textit{overall} (rather than pre-choice) uncertainty, e.g. binning the model's responses into semantic categories and computing the entropy \cite{bakman2024mars, ulmer2024calibrating}. An ensemble-like effect can also be realized by varying the in-context examples provided to the LLM \cite{li2024can}.

\subsubsection{Few-Shot Calibration}
Few-shot calibration techniques employ several queries to the same model and benefit from sequential reasoning as the model evaluates its intermediate generations. For instance, prompting models to begin their responses with a fact and justification for the fact has been shown to improve calibration versus other types of linear reasoning, such as tree-of-thought \cite{zhao2024fact, wei2022chain}. In the domain of code generation, calibration techniques have also been applied to improve the reliability of generated code \cite{spiess2024calibration}. Furthermore, inferring human preferences with in-context learning has been explored as a means to calibrate models in alignment with human judgments \cite{liu2023calibrating}.

\subsubsection{Supervised Calibration}
Supervised calibration approaches, which mainly involve modifying the LLM's weights via additional losses, auxiliary models, or additional data, are also crucial in enhancing model calibration. In supervised methods, learning to classify generated responses as correct (i.e., via a cross-entropy loss) can result in better calibration than non-learning-based approaches and can help to combat overconfidence \cite{chen2022close, zhu2023calibration, johnson2024experts}. In fact, some existing work argue that fine-tuning is necessary for the calibration of uncertainty estimates of LLMs \cite{kapoor2024large}. Given a language generator $\hat f$, score model (confidence) $\hat P$, and a dataset $\mathcal D:=\{(x, y)_i\}_{i=1}^N$ of data-label pairs, the token-level cross-entropy loss seeks to measure the uncertainty of the predicted labels $\hat f(x)$, on average, over the dataset:
${L_{\text{CE}} = -\mathbb E^{(x, y) \sim D} [\log \hat P(y = \hat f(x))],}$
to improve the calibration of the confidence estimates of the model.
While LLMs exhibit high-quality text generations ($\hat f$), their confidences ($\hat P$) may be improved by fine-tuning the model with a cross-entropy loss on the full dataset or a subset. Besides the cross-entropy function, other proper-scoring rules can also be used for achieving calibration \cite{gneiting2007probabilistic, gneiting2007strictly}. Reinforcement learning (with human feedback in some applications) may be used to fine-tune a model to produce realistic confidence estimates, e.g., \cite{band2024linguistic, mao2024don}.
Techniques such as learning to rationalize predictions with generative adversarial networks \cite{sha2021learning}, applying regularization \cite{kong2020calibrated}, and biasing token logits \cite{liu2024litcab, zhao2021calibrate} have also been explored. Finally,  sequence-level likelihood calibration has been proposed to improve the quality of LLM generations \cite{zhao2022calibrating}. 
Instead of modifying the model's weights, another class of supervised calibration methods seeks to modify model \textit{hyperparameters} in a post-hoc manner. These include temperature tuning \cite{desai2020calibration} and methods involving entropy and logit differences [QQ] \cite{lyu2024calibrating}.

\section{Datasets and Benchmarks}
\label{sec:dataset}
Here, we present useful benchmarks in uncertainty quantification for LLMs. The rapid development of highly-capable LLMs has led to the introduction of a slate of benchmarks for measuring advances on the different capabilities of these models. Some examples of these datasets include: GPQA \cite{rein2023gpqa}, a domain-specific dataset with multiple-choice questions in the physical sciences; MMLU \cite{hendrycks2020measuring}, a multi-task dataset for evaluating the breadth of knowledge of LLMs across a wide range of subjects, e.g., the humanities and sciences; HellaSwag \cite{zellers2019hellaswag}, a dataset for evaluating LLM's common-sense reasoning capability in sentence-completion tasks; RACE \cite{lai2017race}, a dataset for reading-comprehension evaluation; GSM8K \cite{cobbe2021training}, a dataset for evaluating the grade-school, math-solving capability of LLMs; and APPS \cite{hendrycks2021measuring}, a code-generation benchmark for LLMs. There have been a related line of work in developing datasets with inherent ambiguities \cite{kamath2024scope, min2020ambigqa, liu2023we, tamkin2022task}, e.g., ``the cat was lost after leaving the house" meaning either that the cat was unable to find the way, or the cat was unable to be found \cite[Fig.~1]{min2020ambigqa}, as well as datasets modeling clarifying questions in multi-turn conversations \cite{aliannejadi2021building}. However, experimental results associated with these datasets do not necessarily incorporate uncertainty evaluation beyond answering accuracy.

\hypertarget{link:read_comp_benchmark}{Although many of the aforementioned benchmarks have not been widely adopted in research on uncertainty quantification, a few benchmarks in natural-language processing have proven highly amenable to research in uncertainty quantification of LLMs, e.g., TriviaQA \cite{joshi2017triviaqa}, a dataset which consists of 95K question-answer pairs for evaluating an LLM's reading-comprehension skill. TriviaQA \cite{joshi2017triviaqa} has been widely utilized in evaluating many methods for uncertainty quantification of LLMs \cite{kuhn2023semantic, mielke2022reducing, stengel2024lacie}. Likewise, other methods have employed CoQA \cite{reddy2019coqa}, a dataset containing conversational question-answer pairs, and WikiBio \cite{lebret2016generating}, a dataset containing biographies from Wikipedia, in evaluating the performance of UQ methods for LLMs.} 
\hypertarget{link:math_benchmark}{The CalibratedMath benchmark was introduced in \cite{lin2022teaching} for examining the ability of LLMs to verbally express their confidence in solving arithmetic tasks.} 
Moreover, datasets for evaluating the consistency of LLMs exist, e.g., ParaRel \cite{elazar2021measuring}, which consists of 328 paraphrases, generated by altering a set of prompts while keeping the semantic meaning of the prompts the same. 
\hypertarget{link:multi_hop_reasoning}{Furthermore, HotpotQA \cite{yang2018hotpotqa} and StrategyQA \cite{geva2021did} represent question-answering benchmarks consisting of question-answer pairs generated from Wikipedia, specifically designed to assess the ability of LLMs to perform multi-hop reasoning.}
\hypertarget{link:factuality_analysis}{Similarly, TruthfulQA \cite{lin2021truthfulqa} represents a factuality-oriented dataset, designed to evaluate the ability of LLMs to generate factual responses to questions that some humans might answer wrongly based on misconceptions. 
Noting the connection between hallucination and uncertainty quantification, uncertain quantification methods can leverage benchmarks for hallucination detection, e.g., HaluEval \cite{li2023halueval}, and datasets for factuality analysis and claim verification, e.g., FEVER \cite{thorne2018fever}.}
Lastly, we note that there has been some work that aims to standardize the tasks for evaluating the performance of LLMs by explicitly accounting for the uncertainty of LLMs in specific tasks, e.g., based on selective classification and generation \cite{vashurin2024benchmarking} or conformal prediction \cite{ye2024benchmarking}.

\section{Applications}
\label{sec:applications}
We highlight a few application areas of uncertainty quantification of LLMs, including its applications to chatbots and other textual use-cases and robotics.

\subsection{Chatbot and Textual Applications}
\label{ssec:chatbot_applications}
Given that LLMs are prone to hallucinate, existing work examines the integration of uncertainty quantification techniques in LLM-enabled chatbots. For example, recent work leverages uncertainty quantification techniques for LLMs in hallucination detection \cite{zhang2023enhancing, yadkori2024believe, kossen2024semantic, tomani2024uncertainty} and content and factuality analysis \cite{tai2024examination, pacchiardi2023catch}. Semantic entropy probes (SEPs) \cite{zhang2023enhancing} utilize linear logistic models to predict semantic entropy from the hidden states of an LLM, demonstrating its effectiveness in detecting hallucinations on a variety of tasks. The approach in \cite{yadkori2024believe} introduces an information-theoretic metric for hallucination detection by estimating both the aleatoric and epistemic uncertainty of the LLM, with the premise that large epistemic uncertainty corresponds to hallucinations. Other downstream applications leverage hallucination detection to estimate the confidence of the LLM on the factuality of its response \cite{mahaut2024factual} or to actively improve the factuality of LLMs during the token-generation step \cite{chang2024real}. 

\begin{figure}[th]
    \centering
    \begin{minipage}[b][][b]{.48\textwidth}
        \centering
        \includegraphics[width=\linewidth]{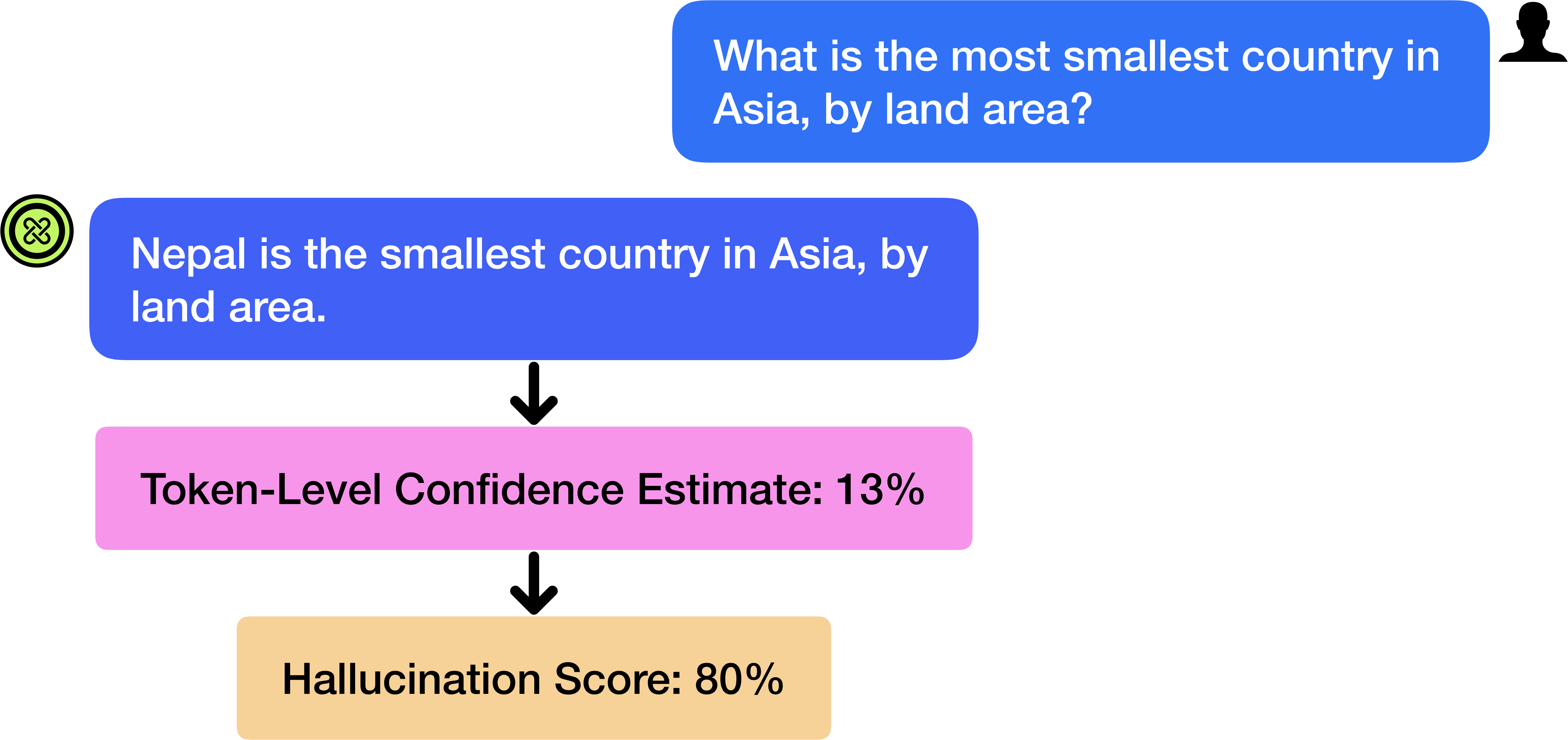}
        \caption{Uncertainty quantification methods for LLMs have been employed in hallucination detection. LLMs tend to be less confident when hallucinating (measured via token-based metrics), although their responses may sound overly confident. In this example, although the LLM provides a confident response to the prompt, a token-level UQ method indicates that the LLM is uncertain, enabling hallucination detection.}
        \Description[Uncertainty quantification methods for LLMs have been employed in hallucination detection. LLMs tend to be less confident when hallucinating (measured via token-based metrics), although their responses may sound overly confident. In this example, although the LLM provides a confident response to the prompt, a token-level UQ method indicates that the LLM is uncertain, enabling hallucination detection.]{Uncertainty quantification methods for LLMs have been employed in hallucination detection. LLMs tend to be less confident when hallucinating (measured via token-based metrics), although their responses may sound overly confident. In this example, although the LLM provides a confident response to the prompt, a token-level UQ method indicates that the LLM is uncertain, enabling hallucination detection.}
        \label{fig:uq_application_hallucination detection}
    \end{minipage}%
    \hfill %
    \begin{minipage}[b][][b]{.48\textwidth}
        \centering
        \includegraphics[width=\columnwidth]{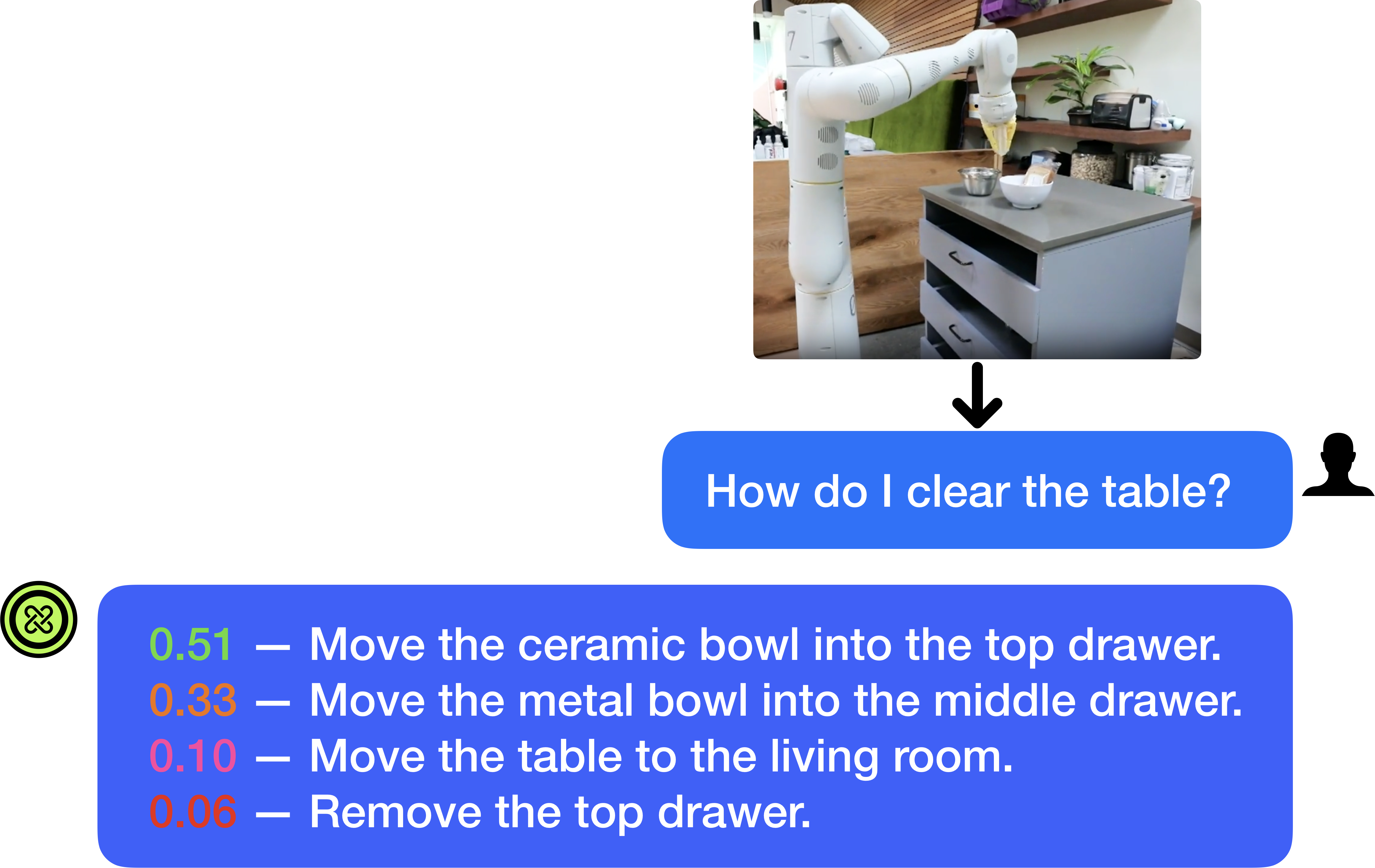}
        \caption{Robotics applications utilize UQ methods to estimate the LLM's confidence in the sub-tasks proposed by the LLM, to determine when human assistance is required.}
        \Description[Robotics applications utilize UQ methods to estimate the LLM's confidence in the sub-tasks proposed by the LLM, to determine when human assistance is required.]{Robotics applications utilize UQ methods to estimate the LLM's confidence in the sub-tasks proposed by the LLM, to determine when human assistance is required.}
        \label{fig:ll_uq_app_robotics}
    \end{minipage}
\end{figure}

In \Cref{fig:uq_application_hallucination detection}, we illustrate an application of uncertainty quantification to detect hallucinations in LLMs. When asked for the smallest country in Asia by land area, the LLM provides a confident response. However, the low token-level confidence estimate reveals the uncertainty of the LLM, indicating a high likelihood of hallucination by the LLM. Drawing upon the association between factuality analysis and uncertainty quantification, the work in \cite{mohri2024language} employs conformal prediction to actively generate outputs that have a high probability of being facts. Further, the work in \cite{pacchiardi2023catch} trains a logistic regression classifier to detect outright lies in LLMs (i.e., false information provided by the LLM when the factual answer is known as opposed to hallucinations where the LLM does not know the factual answer), by asking the LLM follow-up questions unrelated to the original prompt.
Applications in sentiment analysis \cite{maltoudoglou2020bert} and content analysis \cite{xiao2023supporting, dai2023llm, chew2023llm} utilize LLMs in characterizing the sentiments or opinions implied in text sources and in deductive coding to aid the identification of relevant themes across highly-varied documents, respectively. However, noting that LLMs are not necessarily consistent in their outputs, the LLMq method \cite{tai2024examination} examines the LLM's outputs for the presence of epistemic linguistic uncertainty markers and the consistency of the LLM's outputs to identify the thematic codes associated with the text.
Further applications arise in text summarization \cite{kolagar2024aligning}, examining the alignment of uncertainty markers in the original source document and the LLM-generated summary.

Uncertainty quantification has also been explored within the context of \emph{jailbreaking} LLMs. For example, the work in \cite{steindl2024linguistic} examines the connections between predictive entropy and jailbreak prompts, showing that the entropy of the LLM's tokens increases when an LLM is given jailbreak prompts. However, the LLM's uncertainty can be directly manipulated during the jailbreaking attempt \cite{zeng2024uncertainty}. In addition, the evaluation study in \cite{liu2024calibration} highlights that safeguard models for LLMs often show notable miscalibration in jailbreaking attempts.
Further, existing work employs uncertainty quantification techniques to improve LLMs via fine-tuning \cite{osband2022fine, niu2024functional, yang2023improving, yang2024bayesianlowrankadaptationlarge}.
Other applications have explored uncertainty quantification in multi-step interaction and chain-of-thought prompting settings \cite{zhao2024saup, han2024towards}, where the final output of an LLM depends on intermediate responses. To account for the influence of preceding responses, these methods propagate the LLM's uncertainty at each interaction phase. Similar uncertainty propagation techniques have been applied to sequential labeling problems \cite{he2023uncertainty}.
In other applications, uncertainty quantification methods for LLMs have been utilized in retrieval-augmented generation \cite{rouzrokh2024conflare, li2024traq}, using the framework of conformal prediction to provide provable guarantees.
Moreover, some existing work utilizes conformal prediction in response generation from an LLM to identify prediction sets that are likely to contain the ground-truth with some guarantees \cite{quach2023conformal, kumar2023conformal}. 
Although prior work employing conformal prediction generally assume access to the LLM's logits, conformal prediction can also be utilized with black-box LLMs, e.g., \cite{su2024api}.
Lastly, techniques and results from mechanistic interpretability can be used to predict performance of LLMs at test time. In~\cite{schwab2019cxplain}, the authors train a causal explanation model to estimate model performance using sensitivity to input features. In~\cite{nanda2023progress}, the authors find that sudden emergent qualitative changes in LLMs can be predicted by reverse engineering the model.
Further, recent works~\cite{zimmermann2024scale} have shown that scaling up LLMs in terms of model size or dataset does not improve interpretability as previously believed, by surveying human participants.

\subsection{Robotics}
\label{ssec:robotics_applications}
Endowing LLMs with an embodiment (physical form) presents unique challenges, as is the case in robotics. Such embodiment essentially empowers LLMs to be agents of physical change, which can lead to potentially disastrous outcomes if the outputs of the LLMs are not reliable or trustworthy. Although LLMs (and vision-language models) have found widespread applications in robotics, e.g., robotic manipulation \cite{ahn2022can, brohan2022rt, brohan2023rt, kim2024openvla}, robotic navigation and exploration \cite{shah2023lm, dorbala2023can, ren2024explore}, and multi-robot collaboration \cite{kannan2023smart, chen2024scalable, mandi2024roco}, only a few of these applications explicitly consider the uncertainty of the LLMs to ensure safety, although other existing work \cite{wang2024safe} utilize LLMs to assess the success of a task without explicitly reasoning about the confidence of the LLM.

The work in \cite{tsai2024efficient} fine-tunes the Mistral-7B LLM \cite{jiang2023mistral} to generate possible next actions for a decision-making agent and trains a neural point-wise dependency estimator to predict the compatibility score between a user-provided prompt and all generated actions. Subsequently, the authors employ conformal prediction to identify more likely actions for a given prompt, which is presented to the user to select the next action.
A collection of LLM-based task planning work for robots examine the confidence an LLM assigns to its generated next-step plans to determine when human assistance or verification is required, illustrated in \Cref{fig:ll_uq_app_robotics}. 
To determine when an LLM requires clarification from a human, KnowNo \cite{ren2023robots} utilizes a token-based UQ approach to estimate the uncertainty of the LLM in generating possible next steps for a robot given a task, by examining the token probability assigned to each option in the list of possible next steps. Further, KnowNo employs conformal prediction to generate prediction sets over the possible next steps, with provable theoretical guarantees, prompting the human for help, if the prediction set consists of more than one possible action. 
HERACLEs \cite{Wang2023ConformalTL} presents a similar pipeline within a Linear Temporal Logic framework, with multiple high-level sub-goals.

IntroPlan \cite{liang2024introspective} extends KnowNo \cite{ren2023robots} through introspective planning, where, given a task, the LLM retrieves the most relevant instance from a knowledge base constructed from few-shot, human-provided examples and reasons about the feasibility of the possible next actions. Introspective planning enables IntroPlan to generate prediction sets with tighter confidence bounds, minimizing human intervention.
LAP \cite{mullen2024towards} further introduces an action-feasibility metric to improve the alignment of the LLM's confidence estimate with the probability of success, resulting in fewer clarification queries.
S-ATLAS \cite{wang2024safe} extends KnowNo to LLM-based multi-robot task planning, where a team of robot collaborate to complete a task.
In addition, KnowLoop \cite{zheng2024evaluating} utilizes a multi-modal large language model (MLLM), e.g., LLaVa \cite{liu2024visual} or ChatGPT-4V, for failure detection in LLM-based task planning. The MLLM evaluates the success of the task, given images of the environment at each stage, providing its feedback along with its estimated confidence, using either a self-verbalized approach or a token-level UQ method. KnowLoop \cite{zheng2024evaluating} demonstrates that token-level UQ approaches yield better-aligned uncertainty estimates compared to a self-verbalized UQ approach.
Lastly, TrustNavGPT \cite{sun2024trustnavgpt} employs a similar architecture to evaluate the trustworthiness of human commands to an LLM in LLM-based, audio-guided robot navigation.

\section{Open Research Challenges}
\label{sec:open_research_challenges}
We enumerate a number of open research challenges, hoping to drive future research to address these challenges.

\subsection{Consistency is not Factuality}
\label{ssec:challenges_consistency}
\begin{figure}[th]
    \centering
    \begin{minipage}[b][][b]{.48\textwidth}
        \centering
        \includegraphics[width=\columnwidth]{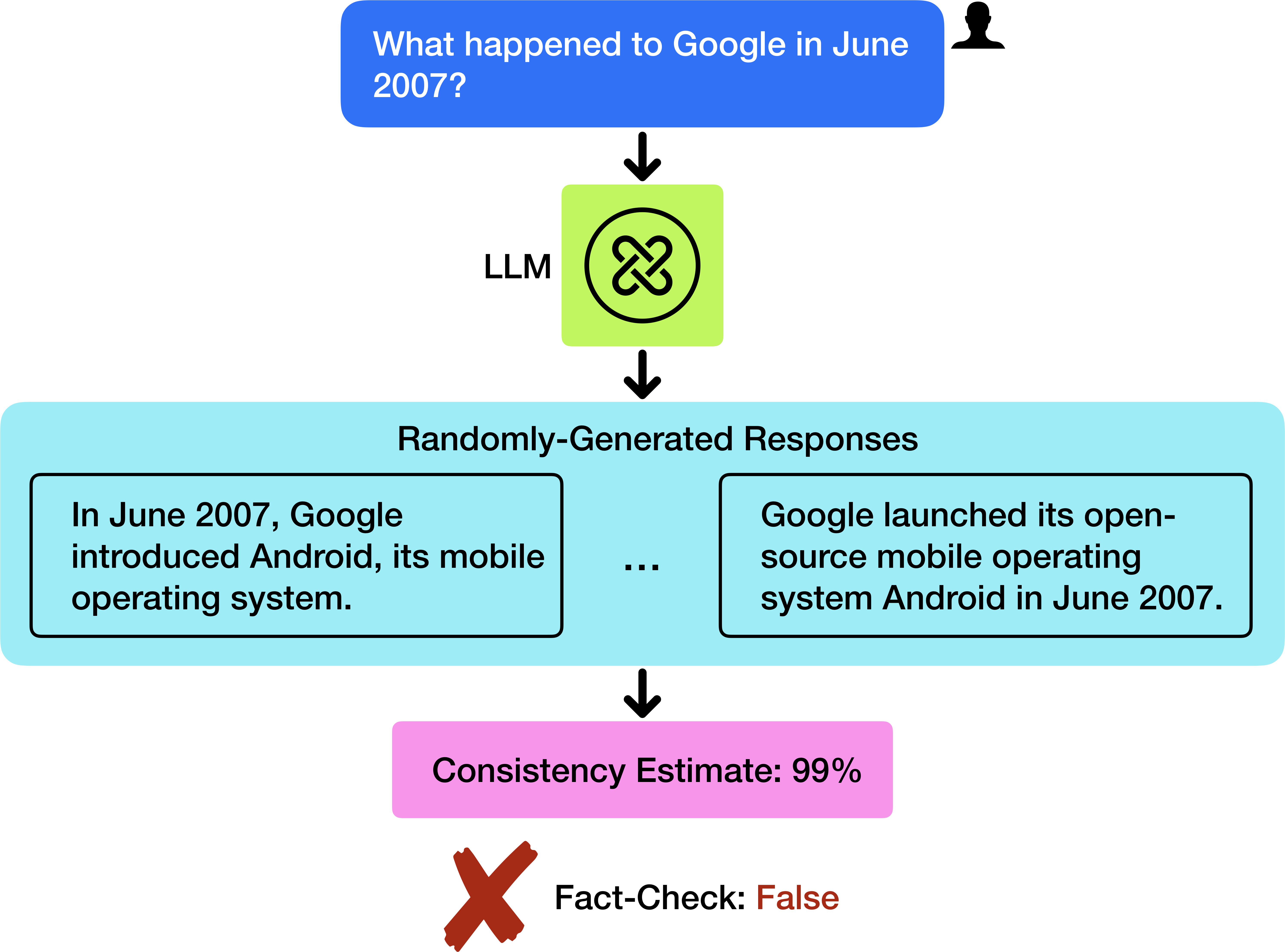}
        \caption{Consistency is not factuality. Semantic-similarity UQ methods for LLMs might provide misleading confidence estimates, e.g., when multiple random responses from the LLM are consistent but false. In this example, the LLM consistently claims that Google introduced Android in June 2007, which is incorrect, given that Android was introduced in November 2007.}
        \Description[Consistency is not factuality. Semantic-similarity UQ methods for LLMs might provide misleading confidence estimates, e.g., when multiple random responses from the LLM are consistent but false. In this example, the LLM consistently claims that Google introduced Android in June 2007, which is incorrect, given that Android was introduced in November 2007.]{Consistency is not factuality. Semantic-similarity UQ methods for LLMs might provide misleading confidence estimates, e.g., when multiple random responses from the LLM are consistent but false. In this example, the LLM consistently claims that Google introduced Android in June 2007, which is incorrect, given that Android was introduced in November 2007.}
        \label{fig:consistency_is_not_factuality}
    \end{minipage}%
    \hfill %
    \begin{minipage}[b][][b]{.48\textwidth}
        \centering
        \includegraphics[width=\columnwidth]{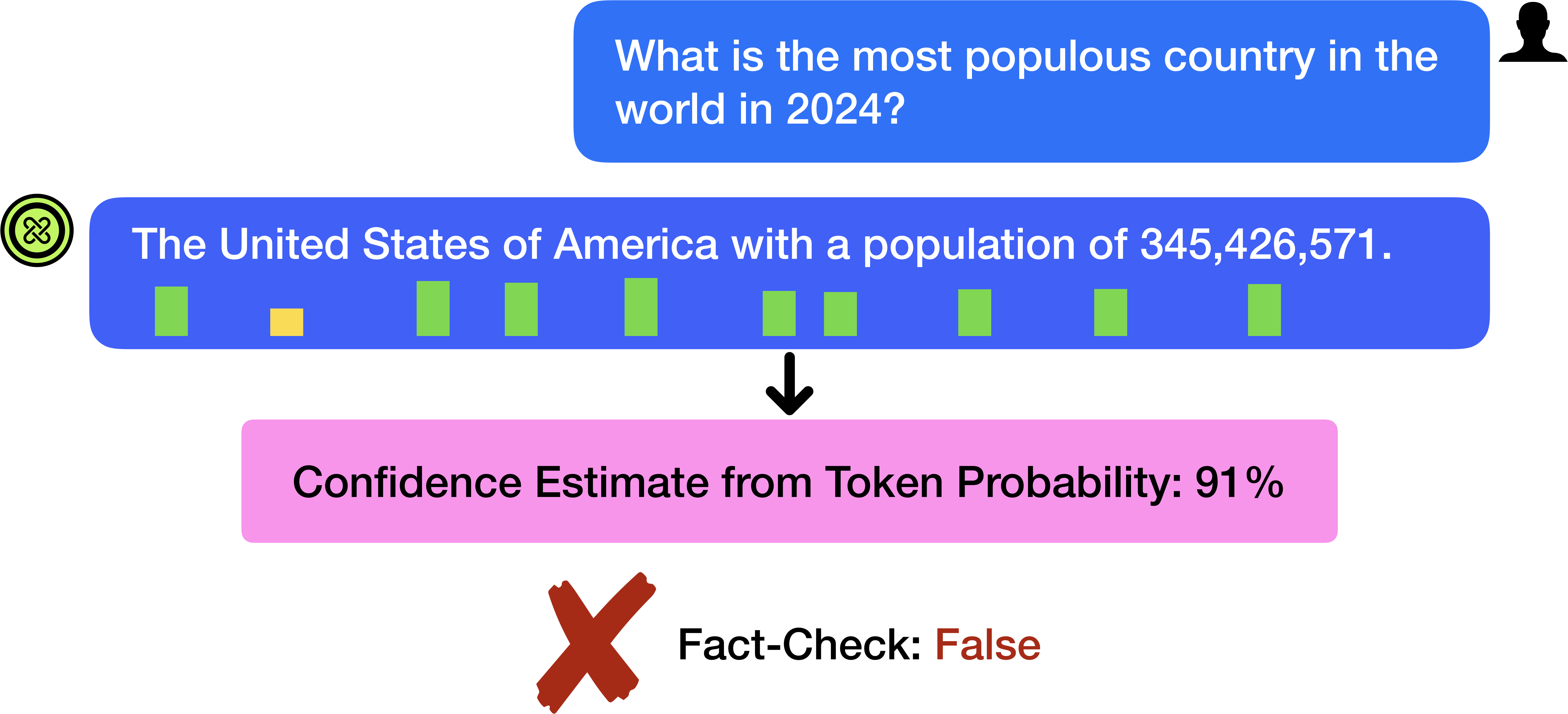}
        \caption{
        Using the conditional distribution of tokens for uncertainty quantification (e.g., in token-level UQ methods) can lead to misleading uncertainty estimates. In this example, the uncertainty of the LLM is notably low, since the succeeding tokens are highly likely given the preceding tokens. However, the claim is incorrect. The most populous country in the world in 2024 is India, not the United States of America. The bars denote the probability of each token.}
        \Description[Using the conditional distribution of tokens for uncertainty quantification (e.g., in token-level UQ methods) can lead to misleading uncertainty estimates. In this example, the uncertainty of the LLM is notably low, since the succeeding tokens are highly likely given the preceding tokens. However, the claim is incorrect. The most populous country in the world in 2024 is India, not the United States of America. The bars denote the probability of each token.]{Using the conditional distribution of tokens for uncertainty quantification (e.g., in token-level UQ methods) can lead to misleading uncertainty estimates. In this example, the uncertainty of the LLM is notably low, since the succeeding tokens are highly likely given the preceding tokens. However, the claim is incorrect. The most populous country in the world in 2024 is India, not the United States of America. The bars denote the probability of each token.}
        \label{fig:entropy_is_not_factuality}
    \end{minipage}
\end{figure}

Many uncertainty quantification methods for LLMs rely on evaluating the consistency between multiple realizations of the response generated by LLMs. This approach faces fundamental limitations, since consistency is not necessarily aligned with factuality. For example, in \Cref{fig:consistency_is_not_factuality}, when prompted to provide a response to the question: ``What happened to Google in June 2007, in a single sentence?" GPT-4 claims that Google announced its mobile operating system Android in June 2007, which is incorrect, given that Android was launched in November 2007. In fact, when creating the set of responses for uncertainty quantification, multiple queries to GPT-4 generate the same incorrect response, which can lead to a miscalibrated confidence estimate. Notably, black-box methods that rely entirely on consistency are most susceptible to this challenge.

Nonetheless, consistency is often a good predictor of factuality, especially when given a sufficiently large number of samples. However, many existing methods do not rigorously examine the number of samples required to define a reliable set of responses when evaluating the consistency of an LLM on a given prompt, which constitutes a critical component for any guarantee on the estimated confidence of the model or factuality of the model's response. Moreover, this challenge might be mitigated by a principled selection of the temperature parameter in an LLM to increase the randomness of the mode; however, the effectiveness of this strategy is quite limited, as excessive randomness in the LLM's outputs defeats the purpose of examining the confidence of the model on a given prompt.

\subsection{Entropy is not Factuality}
\label{ssec:challenges_entropy}
Entropy and other token-based UQ metrics of the token probability distribution in an LLM's output are not necessarily aligned with the factuality of the model's output, although entropy and factuality are often aligned. In particular, the distribution over the tokens is a function of the size of the LLM (including its dictionary of tokens) and the diversity and size of the training data, which can influence the alignment of entropy and factuality. Hence, token-based UQ methods might produce highly miscalibrated confidence estimates for a given prompt, when these estimates are computed entirely from the distribution over the tokens. For example, in a worst-case scenario where the training data is corrupted or insufficient, an LLM might assign most of its probability to an incorrect answer (token) which is most closely related to the training data, leading to a miscalibrated estimate of its confidence.
Moreover, reinforcement learning with human feedback (RLHF), which is utilized in fine-tuning LLMs, generally reduces the calibration of the LLM's confidence estimates \cite{achiam2023gpt}. 
Further, the conditional distribution of each token might not be indicative of the factuality of an LLM's response at the claim-level (sentence-level), i.e., although each generated token might be highly likely given the preceding token, the overall claim expressed by the LLM might not be correct \cite{vazhentsev2024unconditional}, as illustrated in \Cref{fig:entropy_is_not_factuality}.

Future research should explore aligning the entropy of tokens with the factuality of the claims expressed by LLMs and examine augmentation strategies that consider the training distribution of LLMs to better account for the influence of the training data on the probability distribution associated with the generated tokens to ultimately improve the alignment of entropy and other token-based measures of uncertainty with factuality. 
Moreover, the probability distributions over the tokens of an LLM can be manipulated in jailbreaking attacks, leading to misleading confidence estimates and, in some cases, non-factual responses \cite{zeng2024uncertainty}. Future research should seek to improve the robustness of token-level uncertainty quantification methods to adversarial attacks. Further, few existing methods explore uncertainty quantification of LLMs in text summarization, which is critical to the preservation of factual records, constituting an important direction for future research.

\subsection{Applications in Interactive LLM-Enabled Agents}
\label{ssec:challenges_interactive_llm}
Although some existing applications explore uncertainty quantification in LLM-enabled agents, e.g., see \Cref{ssec:robotics_applications}, many of these applications only estimate the LLM's uncertainty at each episode without considering the history of the agent's interaction with the LLM. However, many practical applications require multi-episode interactions, where the LLM generates successive responses based on the information from preceding episodes with the agent. For example, in the scenario depicted in \Cref{fig:ll_uq_app_robotics}, the robot may be asked to prepare a meal for a user, which would require multi-episode interactions, where each episode corresponds to a given sub-task, such as dicing some vegetables before saut\'eing it. Note that utilizing many existing techniques for uncertainty quantification would require the assumption that the LLM's uncertainty at each episode is independent of its prior interaction history, an assumption that is generally not satisfied in real-world applications. Rigorous uncertainty quantification of the LLM's outputs requires the consideration of the history of the agent's interaction with the LLM and its observations (e.g., camera images), in the case of VLMs. This yet-unexplored research area constitutes an exciting direction for future research. 

\subsection{Applications of Mechanistic Interpretability to Uncertainty Quantification}
\label{ssec:challenges_mechanical interpretability}
The connections between interpretability of LLMs and uncertainty quantification have been relatively unexplored, despite the intuitive relationship between both concepts. Mechanistic interpretability holds notable potential in exploiting the synergy between both areas to derive solutions to some of the aforementioned research challenges. For example, the work in \cite{ahdritz2024distinguishing} predicts the token-level confidence of large LLMs using small linear probes (models) trained on the embeddings of frozen pretrained models. This work suggests the existence of a relationship between the internal states of LLMs and their confidence. The authors indicate that their findings suggest that information on the internal state of an LLM could be utilized in distinguishing epistemic uncertainty of the model from aleatoric uncertainty. However, this research area is relatively unexplored, presenting a potentially fruitful direction for future research.

\subsection{Datasets and Benchmarks}
\label{ssec:challenges_datasets_benchmark}
Although a number of datasets and benchmark for uncertainty quantification exists \cite{joshi2017triviaqa, reddy2019coqa, yang2018hotpotqa, lin2021truthfulqa}, to the best of our knowledge, no dataset exists for uncertainty quantification of LLMs in multi-episode interaction scenarios. Future research should examine the creation of versatile, standardized datasets that aid research on uncertainty quantification of LLMs, taking into consideration the history of interaction between a user and an LLM. 
Moreover, benchmarks on uncertainty quantification of LLMs can help inform researchers on the relative performance of their proposed methods. Unfortunately, widely-accepted benchmarks for uncertainty quantification of LLMs do not exist, although some work has been devoted to developing such benchmarks. Future work should seek to create suitable benchmarks for this purpose, especially benchmarks that evaluate the calibration, tightness (conservativeness), and interpretability of uncertainty quantification methods. However, benchmarks can also introduce other challenges, by disconnecting research from practical concerns, overly simplifying the assessment of research advances to outperforming existing work on some metric defined in a benchmark. Hence, care must be taken to ensure that benchmarks remain relevant to practical effectiveness.

\section{Conclusion}
\label{sec:conclusion}
In this survey, we provide a comprehensive review of existing uncertainty quantification methods for LLMs, including relevant background information necessary for readers. We categorize UQ methods for LLMs into four broad classes based on the underlying technique employed by these methods, namely: token-based UQ methods, self-verbalized UQ methods, semantic-similarity-based methods, and mechanistic interpretability. Token-based UQ methods rely on access to an LLM's intermediate outputs or architecture to estimate the confidence an LLM, whereas in self-verbalized UQ methods, the LLM provides its estimated confidence in natural-language. Many semantic-similarity-based methods are black-box methods which only require access to the model's natural-language output, relying on consistency metrics to estimate the LLM's confidence. In contrast, mechanistic interpretability requires access to the LLM's internal activations to identify latent features that explain its activation patterns. Furthermore, we identify relevant datasets and applications for uncertainty quantification of LLMs and highlight open research challenges to inspire future research.

\begin{acks}
We would like to acknowledge Apurva S. Badithela and David Snyder for their contributions. This work was partially supported by the NSF CAREER Award
[\#2044149], the Office of Naval Research [N00014-23-1-2148], and the Sloan Fellowship. Justin Lidard was supported by a National Science Foundation Graduate
Research Fellowship. 
\end{acks}

\bibliographystyle{ACM-Reference-Format}
\bibliography{references}
\end{document}